# Asynchronous Byzantine Machine Learning (the case of SGD)

**Georgios Damaskinos** [1] **El Mahdi El Mhamdi** [1] **Rachid Guerraoui** [1] **Rhicheek Patra** [1] **Mahsa Taziki** [1]

## Abstract

Asynchronous distributed machine learning solutions have proven very effective so far, but always assuming perfectly functioning workers. In practice, some of the workers can however exhibit Byzantine behavior, caused by hardware failures, software bugs, corrupt data, or even malicious attacks. We introduce *Kardam*, the first distributed asynchronous stochastic gradient descent (SGD) algorithm that copes with Byzantine workers. Kardam consists of two complementary components: a filtering and a dampening component. The first is scalar-based and ensures resilience against $\frac{1}{3}$ Byzantine workers. Essentially, this filter leverages the Lipschitzness of cost functions and acts as a self-stabilizer against Byzantine workers that would attempt to corrupt the progress of SGD. The dampening component bounds the convergence rate by adjusting to stale information through a generic gradient weighting scheme. We prove that Kardam guarantees almost sure convergence in the presence of asynchrony and Byzantine behavior, and we derive its convergence rate. We evaluate Kardam on the CIFAR-100 and EMNIST datasets and measure its overhead with respect to non Byzantine-resilient solutions. We empirically show that Kardam does not introduce additional noise to the learning procedure but does induce a slowdown (the cost of Byzantine resilience) that we both theoretically and empirically show to be less than $f/n$, where $f$ is the number of Byzantine failures tolerated and $n$ the total number of workers. Interestingly, we also empirically observe that the dampening component is interesting in its own right for it enables to build an SGD algorithm that outperforms alternative staleness-aware asynchronous competitors in environments with honest workers.

[1]EPFL, Lausanne, Switzerland. Correspondence to: (without space) <firstname.lastname@epfl.ch>.

 

## 1. Introduction

To keep up with the amount of data available today and the corresponding increasing demand for resources, machine learning (ML) practitioners rely on large scale distributed systems (Dean et al., 2012; Abadi et al., 2016; Li et al., 2013; 2014b;a; Ho et al., 2013; Cui et al., 2016). Most of these systems make use of the same work-horse optimization algorithm: *stochastic gradient descent* (SGD), typically following the parameter server scheme (Li et al., 2014a;b). The computation is divided into *epochs*, i.e., *model* (parameter vector) updates. The server gathers gradients from the workers and employs them to perform a single model update, then broadcasts the new model to every worker for computing new gradients (based on random data samples).

To be practical, a distributed ML solution should not assume that all workers perform perfectly well. Some arbitrary behavior of at least a fraction of the workers should be tolerated. The Byzantine failure model (Lamport et al., 1982) offers an elegant abstraction to reason about problems of adversarial machine learning. In particular, a Byzantine behavior can be due to a crash, a software bug, a stale local view of a model, a corrupt piece of data, or worse, to attackers that benefit from a security flaw in a device and compromise its behavior. The Byzantine model encompasses the problem of *poisoning attacks* (Biggio & Laskov, 2012; Muñoz-González et al., 2017; Kurakin et al., 2016). For instance, a group of adversarial (Byzantine) workers could bias the gradient estimator and prevent convergence by sending corrupt gradients. Byzantine-resilient (or simply Byzantine) ML solutions are very appealing for they do not make any assumption on the behavior of Byzantine workers.

A few Byzantine distributed ML solutions have been recently proposed (Blanchard et al., 2017; Su, 2017; El Mhamdi et al., 2018). All however assume a restrictive synchronous model. In each epoch, (1) all (honest) workers are supposed to use the exact same model to compute the gradient, and (2) the parameter server waits for a quorum of workers before aggregating their gradients. When networks are asynchronous, exhibiting heterogeneous (sometimes arbitrary) communication delays, synchronous solutions inevitably lead to slow convergence.

Asynchronous SGD algorithms, on the other hand, enable huge performance benefits despite heterogeneous communi-cation delays (Lian et al., 2016; Liu et al., 2015; Ho et al., 2013; Li et al., 2014a). In short, such algorithms (1) allow workers to make use of a stale model as well as (2) update the model as soon as a new gradient is delivered (instead of waiting for a quorum). Nevertheless, none of these asynchronous algorithms tolerate any Byzantine behavior. In fact, all provably convergent asynchronous SGD algorithms assume that all the workers are permanently honest about their gradient, i.e., provide unbiased estimations of the actual gradient (Figure 1).

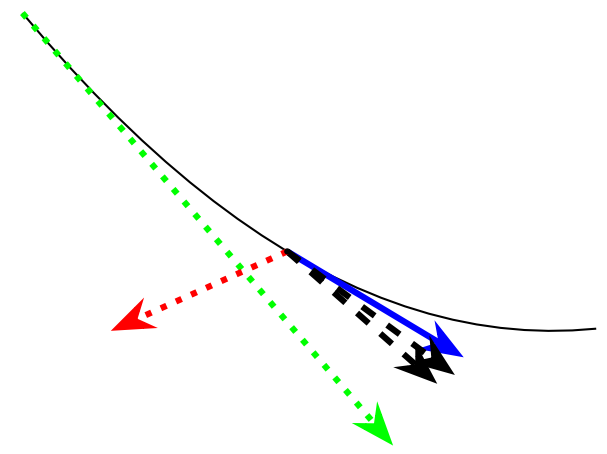

**Figure 1: The gradients computed by non-stale honest workers (black dashed arrows) are distributed around (and are on average equal to) the actual gradient (solid blue arrow) of the cost function (thin black curve). A Byzantine worker can propose an arbitrary poisoning vector (red dotted arrow). A honest but stale worker computes the correct gradient but for a stale version of the model (long green dotted arrow).**

Combining asynchrony with Byzantine resilience is challenging. In particular, aggregating gradients that were computed on different models requires the knowledge of how the curvature of the cost function evolves with staleness. This curvature determines the window of synchrony within which a synchronous method can be transposed into an asynchronous context. Roughly speaking, the more locally curved the cost function is, the narrower this window and vice versa. Estimating the curvature requires heavy computations of the Hessian matrix ($\mathcal{O}(d^2)$), not to mention the fact that this would deprive the parameter server from the most prominent advantage of asynchrony, namely updating the model as soon as a *single* gradient is delivered (i.e., the parameter server would need to aggregate a quorum).

In this paper, we consider for the first time the situation where a significant fraction of workers ($\frac{f}{n}$) can be Byzantine (arbitrarily adversarial) and consider unbounded communication delays. Such situation corresponds to that of many realistic distributed platforms today. We present the first asynchronous Byzantine gradient descent algorithm, we call *Kardam*. Kardam leverages the Lipschitzness of the cost function to filter out gradients from potentially Byzantine workers, while prohibiting Byzantine workers from flooding the parameter server (which in turn would prevent honest workers from updating the model). Kardam also uses a dampening scheme that scales each gradient based on its staleness. The computation overhead for each update is negligible as the filtering component of Kardam is mostly scalar-based. The time complexity for each update computed in terms of the dimension $d$ of a gradient is $\mathcal{O}(d + fn)$. This complexity is the same as the standard complexity of an asynchronous SGD update ($\mathcal{O}(d)$) for the very high-dimensional learning models of today ($d \gg (f, n)$). We prove the convergence of Kardam and precisely determine its convergence rate. In particular, we prove its self-stabilizing property using a refined version of the global confinement argument (Bottou, 1998).

We implemented and deployed Kardam in a distributed setting and we report in this paper on its in-depth empirical evaluation on the CIFAR-100 and EMNIST datasets. In particular, we evaluate the overhead of Kardam with respect to non Byzantine-resilient solutions. Kardam does not tamper with the learning procedure (i.e., include additional noise), yet it does induce a slowdown that we empirically show to be less than $\frac{f}{n}$, where $f$ is the number of Byzantine failures tolerated and $n$ the total number of workers (we also prove that theoretically). Finally, we show that the dampening component (when plugged on an asynchronous non Byzantine-resilient SGD solution) outperforms alternative staleness-aware asynchronous competitors in environments with honest workers.

The code to reproduce our experiments as well as a few additional results (varying $f$) will be found at https://github.com/LPD-EPFL/.

## 2. Computing Model

We consider the general distributed model for machine learning, namely the parameter server (Dean et al., 2012; Abadi et al., 2016; Li et al., 2013; 2014b;a; Ho et al., 2013; Cui et al., 2016)[1]. We assume that $f$ of the $n$ workers are Byzantine (behave arbitrarily). Following the traditional assumption in distributed computing, we assume that the identities of the Byzantine workers are unknown whereas $f$ (in practice, an upper bound) is known. Computation is divided into (infinitely many) asynchronous model updates (epochs).

**Definition 1** (Time)**.** *The global epoch (denoted by $t$) represents the global logical clock of the parameter server (or equivalently the number of model updates). The local timestamp (denoted by $l_p$) for a given worker $p$, represents the epoch of the model that the worker receives from the server and computes the gradient upon. The difference $t - l_p$ can be arbitrarily large due to the asynchrony of the network.*

During each epoch $t$, the parameter server broadcasts the model $\boldsymbol{x}_t \in \mathbb{R}^d$ to all the workers. A cost function $Q$ reflects the quality of the model for the learning task. Each non-Byzantine worker $p$ computes an estimate $\boldsymbol{g}_p = \boldsymbol{G}(\boldsymbol{x}_{l_p}, \xi_p)$ of the actual gradient $\boldsymbol{\nabla} Q(\boldsymbol{x}_{l_p})$ of the cost function $Q$, where $\xi_p$ is a random variable representing, for example, the sample (or a mini-batch of samples) drawn from the dataset at worker $p$. Each worker $p$ sends the timestamp $l_p$ (to declare which version of the model it used) and the gradient $\boldsymbol{g}_p$. See Table 1 for notational details.

[1] Classical techniques of state-machine replication (Lynch, 1996) can be used to ensure that the parameter server is reliable.

| | |
|---|---|
| $t$ | Epoch at the parameter server,incremented after each update. |
| $l_p$ | Timestamp (given by the parameter server) of the model currently used by worker $p$. |
| $\boldsymbol{x}_t$ | Model (parameter vector) at epoch $t$ with dimensionality $d$. |
| $\gamma_t$ | Learning rate at epoch $t$ s.t $\sum_{t=1}^{\infty} \gamma_t = \infty$ and $\sum_{t=1}^{\infty} \gamma_t^2 < \infty$. |
| $\boldsymbol{g}_p$ | Each gradient is a tuple $[\boldsymbol{g}_p, l]$ denoting that a worker $p$ computed the gradient $\boldsymbol{g}_p$ w.r.t $\boldsymbol{x}_l$. |
| $\lvert X \rvert$ | Cardinality of a set $X$. |
| $M$ | Number of gradients that the server waits for before updating the model parameters. ($M = 1$ in asynchrony). |
| $\mathcal{G}_t$ | Set of gradients that the server receives in epoch $t$. Note that $\lvert \mathcal{G}_t \rvert = M$. |
| $\tau_{tl}$ | Staleness value for a gradient $[\boldsymbol{g}, l]$ at epoch $t$ ($\tau_{tl} \triangleq k - l$). |
| $\xi$ | Mini-batch of training examples. |
| $Q(\boldsymbol{x})$ | Cost function for a model $\boldsymbol{x}$. |
| $K$ | Global Lipschitz coefficient of $\boldsymbol{\nabla} Q$, i.e $K = \sup_{\boldsymbol{x},\boldsymbol{y} \in \mathbb{R}^d} \left(\frac{\lVert \boldsymbol{\nabla} Q(\boldsymbol{x}) - \boldsymbol{\nabla} Q(\boldsymbol{y}) \rVert}{\lVert \boldsymbol{x} - \boldsymbol{y} \rVert}\right)$. |

**Table 1: The notations used in Kardam.**

A Byzantine worker $b$ proposes a gradient $\boldsymbol{g}_b$ which can deviate arbitrarily from $\boldsymbol{G}(\boldsymbol{x}_{l_b}, \xi_b)$ (see Figure 1). A Byzantine worker may have full knowledge of the system, including the gradients proposed by other workers. Byzantine workers can furthermore collude, as typically assumed in the distributed computing literature (Lamport et al., 1982; Lynch, 1996; Cachin et al., 2011). Since the communication is assumed to be asynchronous, the parameter server takes into account the first gradient received at time $t$. The parameter server then either suspects the gradient and ignores it, or employs it to update the model and move to epoch $t+1$. We make the following assumptions about any honest worker $p$.

**Assumption 1** (Unbiased gradient estimator)**.**

$$\mathbb{E}_{\xi_p} \boldsymbol{G}(\boldsymbol{x}_{l_p}, \xi_p) = \boldsymbol{\nabla} Q(\boldsymbol{x}_{l_p})$$

**Assumption 2** (Bounded variance)**.**

$$\mathbb{E}_{\xi_p} \lVert \boldsymbol{G}(\boldsymbol{x}_{l_p}, \xi_p) - \boldsymbol{\nabla} Q(\boldsymbol{x}_{l_p}) \rVert^2 \leq d\sigma^2$$

Assumptions 1 and 2 are common in the literature (Bottou, 1998) and hold if the data used for computing the gradients is drawn uniformly and independently.

**Assumption 3** (Linear growth of $r$-th moment)**.**

$$\mathbb{E}_{\xi_p} \lVert \boldsymbol{G}(\boldsymbol{x}, \xi_p) \rVert^r \leq A_r + B_r \lVert \boldsymbol{x} \rVert^r \;\; \forall \boldsymbol{x} \in \mathbb{R}, \;\; r = 2, 3, 4$$

Assumption 3 translates into "the $r$-th moment of the gradient estimator grows linearly with the $r$-th power of the norm of the model" as assumed in (Bottou, 1998).

**Assumption 4** (Lipschitz gradient)**.**

$$\lvert\lvert \boldsymbol{\nabla} Q(\boldsymbol{x}_1) - \boldsymbol{\nabla} Q(\boldsymbol{x}_2) \rvert\rvert \leq K \lvert\lvert \boldsymbol{x}_1 - \boldsymbol{x}_2 \rvert\rvert$$

**Assumption 5** (Convexity in the horizon)**.** *We require that beyond a certain horizon,* $\lVert \boldsymbol{x} \rVert \geq D$*, there exist* $\epsilon > 0$ *and* $0 \leq \beta < \pi/2$ *such that* $\lVert \boldsymbol{\nabla} Q(\boldsymbol{x}) \rVert \geq \epsilon > 0$ *and* $\frac{\langle \boldsymbol{x}, \boldsymbol{\nabla} Q(\boldsymbol{x}) \rangle}{\lVert \boldsymbol{x} \rVert \cdot \lVert \boldsymbol{\nabla} Q(\boldsymbol{x}) \rVert} \geq \cos \beta$.

Assumptions 4 and 5 are the same as in (Blanchard et al., 2017), the first is classic, the second is a slight refinement of a similar assumption in (Bottou, 1998). It essentially states that, beyond a certain horizon $D$ in the parameter space, the opposite of the gradient points towards the origin.

**Definition 2** (Byzantine resilience)**.** *Let Q be any cost function satisfying the assumptions above. Let A be any distributed SGD scheme. We say that A is Byzantine-resilient if the sequence* $\boldsymbol{\nabla} Q(\boldsymbol{x}_t) = 0$ *converges almost surely to zero, despite the presence of up to f Byzantine workers.*

## 3. Kardam

In this section, we present the two main components of our algorithm, *Kardam*[2], namely the filtering and the dampening components. We also establish the theoretical guarantees of each component. For space limitations, the full corresponding proofs are given in the supplementary material.

### 3.1. Byzantine-resilient Filtering Component

The parameter server accepts a gradient $\boldsymbol{g}_p$ from worker $p$ (i.e., updates the model with $\boldsymbol{g}_p$) if $\boldsymbol{g}_p$ is accepted by the Byzantine-resilient filtering component of Kardam. This component itself consists of a *Lipschitz filter* followed by a *frequency filter* that we describe in the following.

**Lipschitz filter.** This filter can be viewed as a kinetic validation at the parameter server based on the empirical Lipschitzness.

**Definition 3** (Empirical Lipschitz coefficient)**.** *The empirical Lipschitz coefficient at worker p is defined as* $\hat{K}_p = \frac{\lVert \boldsymbol{g}_p - \boldsymbol{g}_p^{prev} \rVert}{\lVert \boldsymbol{x}_{l_p} - \boldsymbol{x}_{l_p}^{prev} \rVert}$*. The empirical Lipschitz coefficient at the parameter server is defined with respect to a received gradient from worker p and an updated gradient from worker q at the previous epoch* $(t-1)$ *as* $\hat{K}_t^p = \frac{\lVert \boldsymbol{g}_p - \boldsymbol{g}_q \rVert}{\lVert \boldsymbol{x}_t - \boldsymbol{x}_{t-1} \rVert}$.

The empirical Lipschitz coefficient ($\hat{K}_p$) reflects the local empirical observation of the gradient evolution, normalized by the model evolution. Each worker $p$ derives this coefficient between the current and the previous models used to compute the current and previous gradients of $p$ respectively.

[2]Kardam was a Bulgarian khan who pre-empted the Byzantine empire's invasion. He was the predecessor of *Krum* (Blanchard et al., 2017), the Bulgarian khan gave his name to the first provable solution for the synchronous Byzantine SGD problem.

The Lipschitz filter accepts the candidate gradient $\boldsymbol{g}_p$ if the empirical Lipschitzness for $\boldsymbol{g}_p$ (Definition 3) is not suspicious, i.e., if it is smaller than a median empirical Lipschitzness of all the workers as follows.

$$\hat{K}_t^p \leq \hat{K}_t \triangleq quantile_{\frac{n-f}{n}}\{\hat{K}_p\}_{p\in P}$$

where $quantile_{\frac{n-f}{n}}$ represents the element that separates the $\frac{n-f}{n}$ fraction of workers with the smallest empirical Lipschitz values from the remaining $\frac{f}{n}$ fraction with the highest values (i.e., the $(100 \cdot \frac{n-f}{n})^{th}$ percentile). We highlight that there exist two honest workers $p_1$ and $p_2$ such that $\hat{K}_{p_1} \leq \hat{K}_t \leq \hat{K}_{p_2}$ since our single dimensional median is guaranteed to be bounded by values from any group of size $n-f$ (i.e., group of honest workers).

The complexity of the Lipschitz filter is $\mathcal{O}(d+n)$ (computing distances on 2 $d$-dimensional vectors, then getting the median of $n$ scalars, in $\mathcal{O}(n)$ with quick-select).

Obviously, the Lipschitz filter will end up filtering fast workers (that reach the more curved regions of the cost functions before the others) or slow workers (that are delayed in a curved region while everyone else is already in a less curved region). We note that this filter, roughly speaking, suspects $f$ workers to be Byzantine and thus a pessimistic choice for $f$ would increase the overhead of Kardam (filters more gradients due to a pessimistic choice for $f$).

**Theorem 1** (Optimal Slowdown)**.** *We define the slowdown $SL$ as the ratio between the number of updates from honest workers that pass the Lipschitz filter and the total number of updates delivered at the parameter server. We derive the upper and lower bounds of $SL$ in the following.*

$$\frac{n-2f}{n-f} \leq SL \leq \frac{n-f}{n}$$

*The lower and upper bounds are tight and hold when there are $f$ Byzantine workers and no Byzantine workers respectively. Therefore Kardam achieves the optimal bounds with respect to any Byzantine-resilient SGD scheme and $n \approx 3f$ workers.*

*Proof.* Any Byzantine-resilient SGD scheme assuming $f$ Byzantine workers will at most use $\frac{n-f}{n}$ of the total available workers (upper bound). By definition, the Lipschitz filter accepts the gradients computed by $\frac{n-f}{n}$ of the total workers with empirical Lipschitzness below $\hat{K}_t$. If every worker is honest, then the filter accepts gradients from $\frac{n-f}{n}$ of the workers. We thus get the tightness of the upper bound for the slowdown of Kardam. For the lower bound, the Byzantine workers can know that putting a gradient proposition above $\hat{K}_t$ will get them filtered out and the parameter server will end up using only the honest workers available. The optimal attack would therefore be to slow down the server by getting tiny-Lipschitz gradients accepted while preventing the model from actually changing. This way, the Byzantine workers will make the server filter gradients from a total of $f$ out of the $n-f$ honest workers, leaving only $n-2f$ useful workers for the server. □

**Theorem 2** (Byzantine resilience in asynchrony)**.** *Let $A$ be any distributed SGD scheme. If the maximum successive gradients that $A$ accepts from a single worker and the maximum delay are both unbounded, then $A$ cannot be Byzantine-resilient when $f \geq 1$.*

*Proof.* Without any restrictions, the parameter server could only accept successive gradients from the same Byzantine worker (without getting any update from any honest worker), for example, if the Byzantine worker is faster than any other worker (which is true by the definition of a Byzantine worker and by the fact that delays on (honest) workers are unbounded). This way, the Byzantine worker can force the parameter server to follow arbitrarily bad directions and never converge. Hence, without any restriction on the number of gradients from the workers, we prove the impossibility of asynchronous Byzantine resilience. Readers familiar with distributed computing literature might note that if asynchrony was possible for Byzantine SGD without restricting the number of successive gradients from a single worker, this could be used as an abstraction to solve asynchronous Byzantine consensus (that is impossible to solve (Fischer et al., 1985)). This provides another proof (by contradiction) for our theorem. □

Given Theorem 2 and the objective of making Kardam Byzantine-resilient in an asynchronous environment (i.e., while letting workers be arbitrarily delayed), we introduce the frequency filter.

**Frequency filter.** The goal of this filter is to limit the number of successive gradients [3] from a single worker to a value of $f$, thus not allowing the Byzantine workers to prevent honest workers from updating the model. Consider $L$ as the list of workers who computed the last $2f$ accepted gradients. Assume that the candidate gradient $\boldsymbol{g}_q$ passes the Lipschitz filter. The frequency filter adds worker $q$ at the end of $L$ (i.e., at position $L[2f+1] = q$). If adding this candidate gradient $\boldsymbol{g}_q$ makes any set of workers of size $f$ appear more than $f$ times in $L$, then $q$ is rejected, otherwise, $q$ is accepted. For each worker $p$, the number of times $p$ appeared in $L$ is denoted by $n_p$. The frequency filter accepts a gradient from worker $p$ if the following holds: $\sum_{p\in\theta} n_p \leq f$, where $\theta$ denotes the set of $f$ workers with the $f$ maxima of $\{n_p\}_{p=1}^{n}$. The time complexity of the frequency filter is $\mathcal{O}(2f+1)$ to compute $\{n_p\}_{i=1}^{n}$ (going through the list $L$ of size $2f+1$), in addition to $\mathcal{O}(fn)$ to find the $f$ maxima among $\{n_p\}_{i=1}^{n}$.

[3] One open problem left in our work is the extent to which this filter is too harsh for asynchronous schemes. For instance, it can at least lead to a randomly shuffled round-robin schedule.

**Lemma 1** (Limit of successive gradients)**.** *The frequency filter ensures that any sequence of length $2f+1$ consequently accepted gradients contains at least $f+1$ gradients computed by honest workers.*

*Proof.* Given any sequence of $2f+1$ consequently accepted gradients ($L$), we denote by $S$ the set of workers that computed these gradients. The frequency filter guarantees that any $f$ workers in $S$ computed at most $f$ gradients in $L$. At most $f$ workers in $S$ can be Byzantine, thus at least $f+1$ gradients in $L$ are from honest workers. □

Given asynchrony (unbounded delays), we do not assume any upper bound on the norm of the model, the norm of the gradients or the values of the cost function (regularization schemes can make the loss grow arbitrarily and thereby the gradients norms). However, we assume (as in (Bottou, 1998)) that the cost function is lower bounded by a positive scalar. This assumption holds for all the standard cost functions that are at least lower bounded by zero (e.g., square loss, cross-entropy or any norm-based cost). We denote ***Kar*** by the sequence of gradients accepted (i.e., not filtered) by Kardam, and by $\boldsymbol{Kar}_t$ the gradient accepted by Kardam in epoch $t$.

**Theorem 3** (Correct cone and bounded statistical moments)**.** *If $N > 3f+1$ then for any $t \geq t_r$ (we show that $t_r \in \mathcal{O}(\frac{1}{K\sqrt{|\xi|}})$ where $|\xi|$ is the batch-size of honest workers):*

$$\mathbb{E}(\|\boldsymbol{Kar}_t\|^r) \leq A'_r + B'_r\|\boldsymbol{x}_t\|^r \quad \text{for any r=2, 3, 4 and}$$

$$\langle\mathbb{E}(\boldsymbol{Kar}_t), \boldsymbol{\nabla} Q_t\rangle = \Omega(1 - \frac{\sqrt{d}\sigma}{\|\boldsymbol{\nabla} Q(\boldsymbol{x}_t)\|})\|\boldsymbol{\nabla} Q(\boldsymbol{x}_t)\|^2.$$

*Expectations are on any randomness up to time t conditioned on the past.*

*Proof.* (Sketch) The frequency filter guarantees that there is always an update from a honest worker in any sequence of $f+1$ updates (Lemma 1), i.e., at any time $t$, there is an interval $t-i$ where $i < f+1$ such that the vector that passed the Lipschitz filter is a vector sent by an honest worker (therefore an unbiased estimation of the true gradient). With this in mind, and using triangle inequalities over a series of (at most $f$) previous updates, we prove inequalities on the $r$-th statistical moments of ***Kar***. Those inequalities are in turn plugged into the requirements for the (almost sure) global confinement argument of (Bottou, 1998).

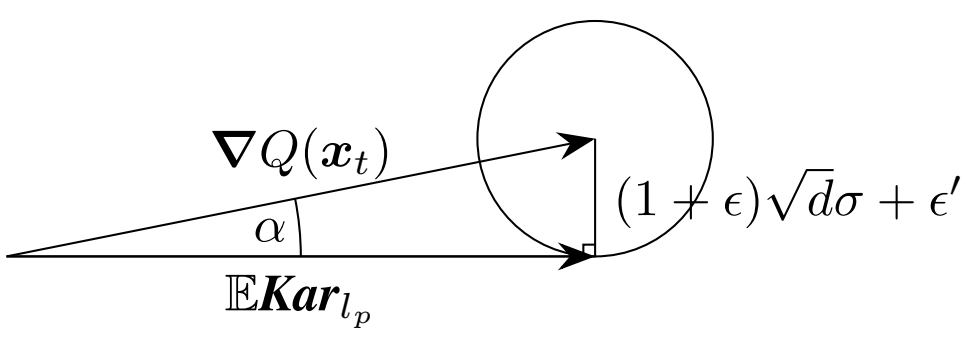

**Figure 2:** If $\|\mathbb{E}\boldsymbol{Kar}_{l_p} - \boldsymbol{\nabla} Q(\boldsymbol{x}_t)\| \leq (1+\epsilon)\sqrt{d}\sigma + \epsilon'$ then $\langle\mathbb{E}\boldsymbol{Kar}_{l_p}, \boldsymbol{\nabla} Q(\boldsymbol{x}_t)\rangle$ is upper bounded by $(1-\sin\alpha)\|\boldsymbol{\nabla} Q(\boldsymbol{x}_t)\|^2$ where $\sin\alpha = \frac{(1+\epsilon)\sqrt{d}\sigma+\epsilon'}{\|\boldsymbol{\nabla} Q(\boldsymbol{x}_t)\|}$.

With the guarantees of almost sure global confinement, and using the Liptschitz properties, and (again) the existence of honest (unbiased) workers in the "recent past" as explained above, we find the lower-bound of the scalar product between the two desired vectors $\langle\mathbb{E}(\boldsymbol{Kar}_t), \boldsymbol{\nabla} Q_t\rangle$ when their distance is small enough compared to their own norms (Figure 2). This finally shows that Kardam remains in a cone of an angle $\alpha$ that is upper bounded by $\arcsin(\frac{(1+\epsilon)\sqrt{d}\sigma+\epsilon'}{\|\boldsymbol{\nabla} Q(\boldsymbol{x}_t)\|})$ with appropriately chosen $\epsilon$ and $\epsilon'$. □

### 3.2. Staleness-aware Dampening Component

We now present the component of Kardam that enables staleness-aware asynchronous updates for the ML model. For the sake of clarity, we denote the time by $t' \triangleq t - tr$ (Theorem 3). We introduce the *M-soft-async* protocol where the server updates the model only after receiving $M$ gradients. The update rule for Kardam is the following.

$$\begin{aligned}\boldsymbol{x}_{t+1} &= \boldsymbol{x}_t - \gamma_t \boldsymbol{Kar}_t \\ &= \boldsymbol{x}_t - \gamma_t \sum_{[\boldsymbol{G}(\boldsymbol{x}_l;\xi_m),l]\in\mathcal{G}_t} \Lambda(\tau_{tl}) \cdot \boldsymbol{G}(\boldsymbol{x}_l;\xi_m) \end{aligned} \quad (1)$$

where $\boldsymbol{G}(\boldsymbol{x}_l;\xi_m)$ denotes the gradient w.r.t the model parameters $\boldsymbol{x}_l$ on the mini-batch $\xi_m$. We assume that every gradient passes the filtering scheme (Section 3.1) at the epoch $t$. Kardam requires $|\mathcal{G}_t| = M$ gradients for each update.

The difference between the standard SGD update rule and our Equation 1 illustrates how Kardam handles asynchronous updates. Kardam dampens each update depending on its staleness value ($\tau_{tl}$). Kardam employs a decay function $\Lambda(\tau_{tl})$ such that $0 \leq \Lambda \leq 1$ to derive the dampening factor for each distinct value of staleness.

**Definition 4** (Dampening function)**.** *We employ a bijective and strictly decreasing dampening function $\tau \mapsto \Lambda(\tau)$ with $\Lambda(0) = 1$.*[4] *Note that every bijective function is also invertible, i.e., $\Lambda^{-1}(\nu)$ exists for every $\nu$ in the range of the $\Lambda$ function.*

Let $\boldsymbol{\Lambda}_t$ be the set of $\Lambda$ values associated with the gradients

[4]If $\Lambda(0) = 1$, then there is no decay for gradients computed on the latest version of the model, i.e., $\tau_{tl} = 0$.

at timestamp $t$.

$$\mathbf{\Lambda}_t = \{\Lambda(\tau_{tl}) \mid [g, l] \in \mathcal{G}_t\}$$

We partition the set $\mathcal{G}_t$ of gradients at timestamp $t$ according to their $\Lambda$-value as follows.

$$\mathcal{G}_t = \bigsqcup_{\lambda \in \mathbf{\Lambda}_t} \mathcal{G}_{t\lambda}$$
$$\mathcal{G}_{t\lambda} = \{[g, l] \in \mathcal{G}_t \mid \Lambda(\tau_{tl}) = \lambda\}$$

Therefore, the update equation can be reformulated as follows.

$$\boldsymbol{x}_{t+1} = \boldsymbol{x}_t - \gamma_t \sum_{\lambda \in \mathbf{\Lambda}_t} \lambda \cdot \sum_{[\boldsymbol{G}(\boldsymbol{x}_l;\xi),l] \in \mathcal{G}_{t\lambda}} \boldsymbol{G}(\boldsymbol{x}_l;\xi)$$

**Definition 5** (Adaptive learning rate). *Given the Lipschitz constant $K$, the total number of timestamps $T$, and the total number of gradients in each timestamp as $M$, we define $\gamma_t$ as follows.*

$$\gamma_t = \underbrace{\sqrt{\frac{Q(\boldsymbol{x}_1) - Q(\boldsymbol{x}^*)}{KTMd\sigma^2}}}_{\gamma} \cdot \underbrace{\frac{M}{\sum_{\lambda \in \mathbf{\Lambda}_t} \lambda |\mathcal{G}_{t\lambda}|}}_{\mu_t} \tag{2}$$

where $\gamma$ is the baseline component of the learning rate and $\mu_t$ is the adaptive component that depends on the amount of stale updates that the server receives at timestamp $t$. Moreover, $\mu_t$ incorporates the total staleness at any timestamp $t$ based on the staleness coefficients ($\lambda$) associated with all the gradients received in timestamp $t$. $Q(\boldsymbol{x}^*)$ (loss value at the optimum) can be assumed to be equal to zero, c.f. supplementary material (Definition 4) for additional comments.

**Remark 1** (Correct cone). *As a consequence of passing the filter and of Theorem 3, $\boldsymbol{G}$ satisfies the following.*

$$\langle \mathbb{E}_\xi \boldsymbol{G}(\boldsymbol{x};\xi), \boldsymbol{\nabla} Q(\boldsymbol{x}) \rangle > \Omega((\|\boldsymbol{\nabla} Q(x_t)\| - \sqrt{d}\sigma)\|\boldsymbol{\nabla} Q(x_t)\|)$$

The theoretical guarantee for the convergence rate of Kardam depends on Assumptions 2,4 and Remark 1. These assumptions are weaker than the assumptions for the convergence guarantees in (Zhang et al., 2016b; Jiang et al., 2017). In particular, due to unbounded delays and the potential presence of Byzantine workers, we only assume the unbiased gradient estimator $\boldsymbol{G}(\cdot)$ for honest workers (Assumption 1). We instead employ (Remark 1) the fact that $\boldsymbol{G}(\cdot)$ and $\boldsymbol{\nabla} Q(\boldsymbol{x})$ make a lower bounded angle together (and subsequently a lower bounded scalar product) for all the workers. The classical unbiased assumption is more restrictive as it requires this angle to be exactly equal to 0, and the scalar product to be equal to $\|\boldsymbol{\nabla} Q(\boldsymbol{x})\| \cdot \|\boldsymbol{G}(\boldsymbol{x})\|$. Most importantly, we highlight the fact that those assumptions are satisfied by Kardam, since every gradient used in this section to compute the ***Kar*** update has passed the Lipschitz filter of the previous section.

**Theorem 4** (Convergence guarantee). *We express the convergence guarantee in terms of the ergodic convergence, i.e., the weighted average of the $\mathcal{L}_2$ norm of all gradients ($||\nabla Q(\boldsymbol{x}_t)||^2$). Using the above-mentioned assumptions, and the maximum adaptive rate $\mu_{\max} = \max\{\mu_1, \ldots, \mu_t\}$, we get the following bound on the convergence rate.*

$$\frac{1}{T}\sum_{t=1}^{T} \mathbb{E}||\nabla Q(\boldsymbol{x}_t)||^2 \leq (2 + \mu_{\max} + \gamma K M \chi \mu_{\max})\, \gamma K d\sigma^2 + d\sigma^2 + 2DK\sigma\sqrt{d} + K^2 D^2$$

*under the prerequisite that*

$$\sum_{\lambda \in \mathbf{\Lambda}_t} \lambda^2 |\mathbf{\Lambda}_t| \left\{ K\gamma_t^2 + \sum_{s=1}^{\infty} \left( \sum_{\nu \in \mathbf{\Lambda}_{t+s}} \gamma_{t+s} K^2 \nu |\mathcal{G}_{t+s,\nu}| \Lambda^{-1}(\nu) \mathbb{I}_{(s \leq \Lambda^{-1}(\nu))} \gamma_t^2 \right) \right\} \leq \sum_{\lambda \in \mathbf{\Lambda}_t} \frac{\gamma_t \lambda}{|\mathcal{G}_{t\lambda}|} \tag{3}$$

*where the Iverson indicator function is defined as follows.*

$$\mathbb{I}_{(s \leq \Delta)} = \begin{cases} 1 & \text{if } s \leq \Delta \\ 0 & \text{otherwise.} \end{cases}$$

It is important to note that the prerequisite (Inequality 3) holds for any decay function $\Lambda$ (since $\lambda < 1$ holds by definition) and for any standard learning rate schedule such that $\gamma_t < 1$. Various SGD approaches (Zhang et al., 2016b; Lian et al., 2015; Zhang et al., 2016a; Jiang et al., 2017) provide convergence guarantees with similar prerequisites.

**Theorem 5** (Convergence time complexity). *Given any mini-batch size $|\xi|$, the number of gradients $M$ the server waits for before updating the model, and the total number of epochs $T$, the time complexity for the convergence of Kardam is:*

$$\mathcal{O}\left(\frac{\mu_{\max}}{\sqrt{T|\xi|M}} + \frac{\chi\mu_{\max}}{T} + d\sigma^2 + 2DK\sigma\sqrt{d} + K^2D^2\right)$$

*where $\chi$ denotes a constant such that for all $\tau_{tl}$, the following inequality holds:*

$$\tau_{tl} \cdot \Lambda(\tau_{tl}) \leq \chi \tag{4}$$

Theorem 5 highlights the relation between the staleness and the convergence time complexity. This time complexity is linearly dependent on the decay bound ($\chi$) and the maximum adaptive rate ($\mu_{max}$).

**Remark 2** (Dampening comparison). *Given two dampening functions $\Lambda_1(\tau) = \frac{1}{1+\tau}$ and $\Lambda_2(\tau) = exp(-\alpha\sqrt[\beta]{\tau})$, and the convergence time complexity from Theorem 5, $\Lambda_2(\tau)$ converges faster than $\Lambda_1(\tau)$ when $\frac{\beta}{e} < \alpha \leq \frac{ln(\tau+1)}{\sqrt[\beta]{\tau}}$.*

We also empirically highlight Remark 2 by comparing these two functions in Figure 4 where DynSGD (Jiang et al., 2017) employs $\Lambda_1$ and Kardam employs $\Lambda_2$.

## 4. Experiments

In this section, we report on our empirical evaluation of our distributed implementation of Kardam. Experiments on Byzantine attacks are mostly illustrative for (1) the importance of the dampening component and (2) the overhead of the filtering component. Due to the intractability of testing all possible attacks, the only option is to prove Byzantine resilience mathematically and focus in the empirical part on the performance overhead of Kardam.

We employ the convolutional neural network (CNN) described in Table 2 for image classification on CIFAR-100. The chosen base learning rate is $15 * 10^{-4}$ and the mini-batch size is 100 examples (Neyshabur et al., 2015). If not stated otherwise, we employ a setup with no actual Byzantine behavior and deploy Kardam with $f = 3$ on 10 workers.

| Parameters | Input | Conv1 | Pool1 | Conv2 | Pool2 | FC1 | FC2 | FC3 |
|---|---|---|---|---|---|---|---|---|
| Kernel size | 32×32×3 | 3×3×16 | 3×3 | 3×3×64 | 4×4 | 384 | 192 | 100 |
| Strides | | 1×1 | 3×3 | 1×1 | 4×4 | | | |

**Table 2: CNN parameters for CIFAR-100.**

**Staleness-aware learning.** We simulate a Gaussian staleness distribution (Zhang et al., 2016b) and evaluate Kardam with different dampening functions $\Lambda(\tau)$ (Definition 4) shown in Figure 3(a). We compare with the performance of Kardam without the Byzantine resilience component (BASELINE-ASGD) by using the constant function ($\Lambda_1 = 1$). Additionally, we compare with a state-of-the-art staleness-aware learning algorithm (DYNSGD (Jiang et al., 2017)) that employs an inverse dampening function ($\Lambda_2(\tau) = \frac{1}{1+\tau}$). Finally, we use two exponential functions ($\Lambda = exp(-\alpha * \tau)$) which, to the best of our knowledge, only Kardam enables.

Figure 3 depicts the very fact that the staleness-aware component of Kardam is crucial in asynchronous environments. We simulate a Gaussian distribution (Figure 3(b)) (similar to (Zhang et al., 2016b)) and show that SSGD has the faster convergence whereas BASELINE-ASGD diverges (Figures 3(b) and 3(c)).

Figure 3 also highlights the need for an adjustable smoothness on the dampening function. A very steep function ($\Lambda_2$) almost ignores many of the updates (weighted by a very small value) and thus suffers a slower convergence. A tuned exponential function ($\Lambda_3$) accelerates the convergence in comparison with the inverse function of DYNSGD. Moreover, Kardam ($\Lambda_3$) assigns larger weights to the less stale updates ($\tau < 13$) compared to DYNSGD and vice versa.

The dampening function selection is the outcome of adjusting the trade-off between the robustness and the magnitude of each update. We observe similar results for the EMNIST dataset (in our supplementary material) and thus highlight that the dampening function can be selected based on the expected staleness distribution, and not necessarily adjusted for each different application.

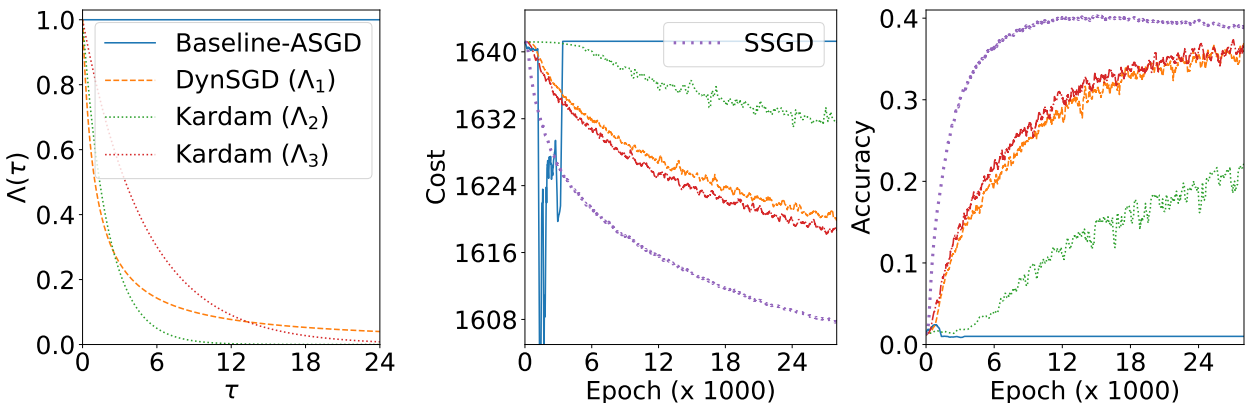

(a) Dampening functions (b) Cost (c) Accuracy

**Figure 3: Staleness-aware learning for CIFAR-100. BASELINE-ASGD denotes Kardam without the dampening component and SSGD the ideal (synchronous) SGD execution. The staleness follows a Gaussian distribution ($mean = 12$, $\sigma = 4$) and the dampening functions are $\Lambda_1 = \frac{1}{\tau+1}$, $\Lambda_2 = e^{0.5\tau}$, $\Lambda_3 = e^{0.2\tau}$.**

**Impact of staleness.** An increase in the amount of staleness leads to a slower convergence according to Theorem 5 (i.e., larger $\chi$ in Equation 4). Figure 4 depicts the impact of the amount of staleness on Kardam and DYNSGD for two different staleness distributions ($D_1$ and $D_2$). We observe that the smaller the mean of the distribution, the faster the convergence. We verify that Kardam outperforms DYNSGD for $D_1$ and vice versa for $D_2$, as the values of the exponential dampening function become very small for the larger staleness values ($D_2$). We highlight that our experimental setup includes significantly higher staleness (D2) than the competitors (Zhang et al., 2016b; Jiang et al., 2017).

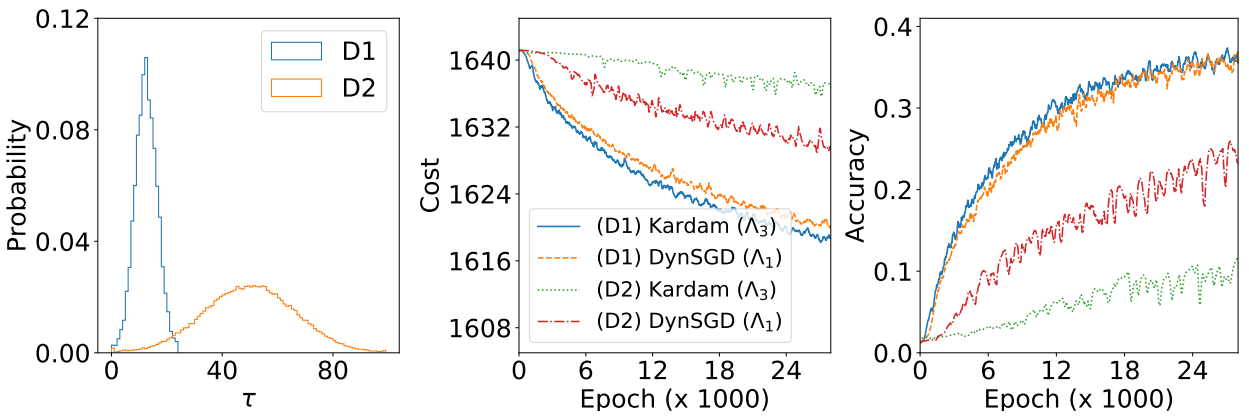

(a) Staleness distributions (b) Cost (c) Accuracy

**Figure 4: Impact of staleness for CIFAR-100.**

**Byzantine resilience.** We observe that the overhead of the Byzantine resilience in the setup with no actual Byzantine behavior is only in terms of filtered (i.e., wasted) gradients and not in terms of convergence speed (in terms of epochs). Moreover, the drop ratio under the staleness distribution $D_1$ is 27.9% and 19.6% for Kardam employing $\Lambda_1$ and $\Lambda_3$ respectively, thus aligned with our theoretical bound (Theorem 1). The slowdown would decrease accordingly by decreasing $f$, i.e., being more optimistic about the number of Byzantine workers.

We test Kardam against a *baseline* Byzantine behavior (3 out of 10 workers send $g_p^{byz} = -10 g_p$) and observe that Kardam successfully filters 100% of the Byzantine gradients (an empirical confirmation of the theoretically proven Byzantine resilience of Kardam).

## 5. Related Work and Concluding Remarks

Kardam is, to the best of our knowledge, the first asynchronous distributed SGD algorithm that tolerates Byzantine behavior. In the following, we discuss papers that either address asynchrony or Byzantine behavior.

**Asynchronous stochastic gradient descent.** SGD is used widely in ML solutions due to its convergence guarantees with low time complexity per update, low memory cost, and robustness against noisy gradients. Several variants of SGD have been proposed to improve the convergence rate and the robustness against noise. Stale-synchronous parallel (Ho et al., 2013; Cui et al., 2014) or bulk-synchronous (Zinkevich et al., 2010; Chen et al., 2016) variants typically target settings with limited staleness due to the limited performance variability among the computing devices. Other approaches consider variance minimization by importance sampling (Alain et al., 2015). The theoretical guarantees underlying these approaches assume synchronous updates as well as a specific formula to compute a gradient norm on each sample, which is only valid for multilayer perceptrons. The scheduler in (Zhang et al., 2016a) assumes all workers to be constantly available, which makes the algorithm not applicable to our setting with Byzantine workers. (Jiang et al., 2017) recently introduced a stale-synchronous parallel (SSP) heterogeneity-aware algorithm. SSP algorithms assume bounded staleness while Kardam guarantees convergence without any such bound (i.e., asynchronous parallel). Additionally, Kardam provides the flexibility of choosing the appropriate dampening function according to the expected staleness distribution while being Byzantine tolerant and asynchronous. We show both theoretically (Remark 2) and empirically (Figure 3) that an exponential dampening function leads to a faster convergence. (Kaoudi et al., 2017) recently proposed an elegant optimizer to predict the optimal SGD variant based on the expected cost per iteration and the estimated number of iterations. This estimation does not however account for stale updates. Our convergence analysis for Kardam could be employed to estimate the number of iterations for different dampening functions and hence to predict the optimal staleness-aware SGD variant. (Lock-freedom) Kardam, put *before* lock-free solutions such as Hogwild (Recht et al., 2011) would not break the convergence requirements (since the purpose of Kardam is to preserve them despite Byzantine workers). However, Kardam and Hogwild do not commute.

**Second order methods.** These methods rely on computing the Hessian matrix instead of the Lipschitz factor (Kardam filtering component). They were not specifically designed for Byzantine resilience but can in fact be employed for that purpose. However, unlike our scalar-based Lipschitz filter ($\mathcal{O}(d)$ time complexity that is already within the usual cost of an SGD update), they suffer from the curse of dimensionality. Moreover, the parameter server does no less than $\Omega(d^2)$ verifications on the Hessian matrix or on the gradient covariance matrix. In the presence of a cheap (constant size $K$) heuristic, the parameter server will let the Byzantine worker with a margin of $d^2 - K$ open coordinates to use for an attack. Since $d \gg K$ the heuristic alternative clearly hampers Byzantine resilience.

**The differentiability lenses of Lipschitz.** A central piece of our work is to filter out suspected vectors based on their (lack of) similar Lipschitzness with the median behavior. We prove that this *filtering* idea is sound, given that a significant fraction ($\Omega(\frac{n-f}{n})$) of workers will almost surely pass it and that Byzantine workers passing it are not harmful. In fact, leveraging the Lipschitzness properties, in the differentiable context of gradient-based learning, is not an uncommon idea. It was used in different contexts, for example, to understand fine-grained robustness, i.e robustness of the model to internal errors at the level of neurons and/or weights, this was done in (El Mhamdi & Guerraoui, 2016; 2017; El Mhamdi et al., 2017) proving a tight upper bound on the Lipschitz coefficient of neural networks, and deriving an exponential dependency with the depth and a polynomial dependency with the Lipschitz coefficient of the activation function used in each layer. In the same time, Lipschitzness was leveraged to compute spectral bounds as in (Cisse et al., 2017; Bartlett et al., 2017) both of which observed the same exponential dependency on the depth. In fact, manipulating differentiable objects is what makes the world of learning fundamentally different from the usual world of distributed computing, where the focus is on combinatorial and discrete structures. The differentiability of learning algorithms acts as a source of relaxation to solve a distributed computing task (estimating a gradient, distributively) in asynchrony and in the presence of Byzantine workers. The shorter the time it takes for Kardam to self-stabilize ($t_r$) the better in term of the speed of convergence. As we prove in Theorem 3, $t_r$ is shorter with a larger global Lipschitz coefficient, i.e., steeper cost functions. Nevertheless, the cost function cannot be controlled. Yet, $t_r$ can be decreased by increasing the batch size per worker, which is no surprise in learning theory (increasing the batch size is one of the most unavoidable taxes (Blanchard et al., 2017; Bousquet & Bottou, 2008; El Mhamdi et al., 2018) for increasing robustness). In practice, our experiments show no significant impact from $t_r$ in the absence of actual Byzantine workers. In their presence, Kardam remains, to the best of our knowledge, the first provably Byzantine-resilient option to run SGD asynchronously.

An open problem now is how to tackle the Byzantine question in asynchronous machine learning *beyond* gradient-based algorithms. We argue that the core idea we present –filtering on quantiles from the recent past– could have applications to any approach where updates arrive with suspicions on either staleness or malicious behavior.

## References

Abadi, Martín, Barham, Paul, Chen, Jianmin, Chen, Zhifeng, Davis, Andy, Dean, Jeffrey, Devin, Matthieu, Ghemawat, Sanjay, Irving, Geoffrey, Isard, Michael, et al. Tensorflow: A system for large-scale machine learning. In *OSDI*, 2016.

Alain, Guillaume, Lamb, Alex, Sankar, Chinnadhurai, Courville, Aaron, and Bengio, Yoshua. Variance reduction in sgd by distributed importance sampling. *arXiv preprint arXiv:1511.06481*, 2015.

Bartlett, Peter L, Foster, Dylan J, and Telgarsky, Matus J. Spectrally-normalized margin bounds for neural networks. In *NIPS*, pp. 6241–6250, 2017.

Biggio, Battista and Laskov, Pavel. Poisoning attacks against support vector machines. In *ICML*, 2012.

Blanchard, Peva, El Mhamdi, El Mahdi, Guerraoui, Rachid, and Stainer, Julien. Machine learning with adversaries: Byzantine tolerant gradient descent. In *NIPS*, pp. 118–128, 2017.

Bottou, Léon. Online learning and stochastic approximations. *Online learning in neural networks*, 17(9):142, 1998.

Bousquet, Olivier and Bottou, Léon. The tradeoffs of large scale learning. In *NIPS*, pp. 161–168, 2008.

Cachin, Christian, Guerraoui, Rachid, and Rodrigues, Luis. *Introduction to reliable and secure distributed programming*. 2011.

Chen, Jianmin, Monga, Rajat, Bengio, Samy, and Jozefowicz, Rafal. Revisiting distributed synchronous sgd. *arXiv preprint arXiv:1604.00981*, 2016.

Cisse, Moustapha, Bojanowski, Piotr, Grave, Edouard, Dauphin, Yann, and Usunier, Nicolas. Parseval networks: Improving robustness to adversarial examples. In *ICML*, pp. 854–863, 2017.

Cui, Henggang, Cipar, James, Ho, Qirong, Kim, Jin Kyu, Lee, Seunghak, Kumar, Abhimanu, Wei, Jinliang, Dai, Wei, Ganger, Gregory R, Gibbons, Phillip B, et al. Exploiting bounded staleness to speed up big data analytics. In *USENIX ATC*, pp. 37–48, 2014.

Cui, Henggang, Zhang, Hao, Ganger, Gregory R, Gibbons, Phillip B, and Xing, Eric P. Geeps: Scalable deep learning on distributed gpus with a gpu-specialized parameter server. In *European Conference on Computer Systems*, pp. 4, 2016.

Dean, Jeffrey, Corrado, Greg, Monga, Rajat, Chen, Kai, Devin, Matthieu, Mao, Mark, Senior, Andrew, Tucker, Paul, Yang, Ke, Le, Quoc V, et al. Large scale distributed deep networks. In *NIPS*, pp. 1223–1231, 2012.

El Mhamdi, E. M. and Guerraoui, R. When neurons fail. In *IPDPS*, pp. 1028–1037, May 2017.

El Mhamdi, E. M., Guerraoui, R., and Rouault, S. On the robustness of a neural network. In *SRDS*, pp. 84–93, Sept 2017.

El Mhamdi, El Mahdi and Guerraoui, Rachid. When neurons fail - technical report. pp. 19, 2016. Biological Distributed Algorithms Workshop, Chicago.

El Mhamdi, El Mahdi, Guerraoui, Rachid, and Rouault, Sébastien. The hidden vulnerability of distributed learning in byzantium. *arXiv preprint arXiv:1802.07927*, 2018.

Fischer, Michael J, Lynch, Nancy A, and Paterson, Michael S. Impossibility of distributed consensus with one faulty process. *JACM*, 32(2):374–382, 1985.

Ho, Qirong, Cipar, James, Cui, Henggang, Lee, Seunghak, Kim, Jin Kyu, Gibbons, Phillip B, Gibson, Garth A, Ganger, Greg, and Xing, Eric P. More effective distributed ml via a stale synchronous parallel parameter server. In *NIPS*, pp. 1223–1231, 2013.

Jiang, Jiawei, Cui, Bin, Zhang, Ce, and Yu, Lele. Heterogeneity-aware distributed parameter servers. In *SIGMOD*, pp. 463–478, 2017.

Kaoudi, Zoi, Quiané-Ruiz, Jorge-Arnulfo, Thirumuruganathan, Saravanan, Chawla, Sanjay, and Agrawal, Divy. A cost-based optimizer for gradient descent optimization. In *SIGMOD*, pp. 977–992, 2017.

Kurakin, Alexey, Goodfellow, Ian, and Bengio, Samy. Adversarial machine learning at scale. *arXiv preprint arXiv:1611.01236*, 2016.

Lamport, Leslie, Shostak, Robert, and Pease, Marshall. The byzantine generals problem. *TOPLAS*, 4(3):382–401, 1982.

Li, Mu, Zhou, Li, Yang, Zichao, Li, Aaron, Xia, Fei, Andersen, David G, and Smola, Alexander. Parameter server for distributed machine learning. In *Big Learning NIPS Workshop*, volume 6, pp. 2, 2013.

Li, Mu, Andersen, David G, Park, Jun Woo, Smola, Alexander J, Ahmed, Amr, Josifovski, Vanja, Long, James, Shekita, Eugene J, and Su, Bor-Yiing. Scaling distributed machine learning with the parameter server. In *OSDI*, volume 1, pp. 3, 2014a.

Li, Mu, Andersen, David G, Smola, Alexander J, and Yu, Kai. Communication efficient distributed machine learning with the parameter server. In *NIPS*, pp. 19–27, 2014b.

Lian, Xiangru, Huang, Yijun, Li, Yuncheng, and Liu, Ji. Asynchronous parallel stochastic gradient for nonconvex optimization. In *NIPS*, pp. 2737–2745, 2015.

Lian, Xiangru, Zhang, Huan, Hsieh, Cho-Jui, Huang, Yijun, and Liu, Ji. A comprehensive linear speedup analysis for asynchronous stochastic parallel optimization from zeroth-order to first-order. In *NIPS*, 2016.

Liu, Ji, Stephen, and Wright, J. Asynchronous stochastic coordinate descent: Parallelism and convergence properties. Technical report, SIAM Journal on Optimization, 2015.

Lynch, Nancy A. *Distributed algorithms*. 1996.

Muñoz-González, Luis, Biggio, Battista, Demontis, Ambra, Paudice, Andrea, Wongrassamee, Vasin, Lupu, Emil C, and Roli, Fabio. Towards poisoning of deep learning algorithms with back-gradient optimization. In *Workshop on Artificial Intelligence and Security*, pp. 27–38, 2017.

Neyshabur, Behnam, Salakhutdinov, Ruslan R, and Srebro, Nati. Path-sgd: Path-normalized optimization in deep neural networks. In *NIPS*, pp. 2422–2430, 2015.

Recht, Benjamin, Re, Christopher, Wright, Stephen, and Niu, Feng. Hogwild: A lock-free approach to parallelizing stochastic gradient descent. In *NIPS*, pp. 693–701, 2011.

Su, Lili. *Defending distributed systems against adversarial attacks: consensus, consensus-based learning, and statistical learning*. PhD thesis, University of Illinois at Urbana-Champaign, 2017.

Zhang, Ruiliang, Zheng, Shuai, and Kwok, James T. Asynchronous distributed semi-stochastic gradient optimization. In *AAAI*, pp. 2323–2329, 2016a.

Zhang, Wei, Gupta, Suyog, Lian, Xiangru, and Liu, Ji. Staleness-aware async-sgd for distributed deep learning. In *IJCAI*, pp. 2350–2356, 2016b.

Zinkevich, Martin, Weimer, Markus, Li, Lihong, and Smola, Alex J. Parallelized stochastic gradient descent. In *NIPS*, pp. 2595–2603, 2010.

# Asynchronous Byzantine Machine Learning (the case of SGD) Supplementary Material

**Abstract**

We theoretically prove the Byzantine resilience (Section 1) and convergence (Section 2) of Kardam. Furthermore, we provide additional experimental results for the EMNIST dataset in Section 3.

## 1 Analysis of Byzantine Resilience

**Definition 1** (Time). *The global epoch (denoted by $t$) represents the global logical clock of the parameter server (or equivalently the number of model updates). The local timestamp (denoted by $l_p$) for a given worker $p$, represents the epoch of the model that the worker receives from the server and computes the gradient upon. The difference $t - l_p$ can be arbitrarily large due to the asynchrony of the network.*

We make the following assumptions about any honest worker $p$.

**Assumption 1** (Unbiased gradient estimator).

$$\mathbb{E}_{\xi_p} \boldsymbol{G}(\boldsymbol{x}_{l_p}, \xi_p) = \boldsymbol{\nabla} Q(\boldsymbol{x}_{l_p})$$

**Assumption 2** (Bounded variance).

$$\mathbb{E}_{\xi_p} \|\boldsymbol{G}(\boldsymbol{x}_{l_p}, \xi_p) - \boldsymbol{\nabla} Q(\boldsymbol{x}_{l_p})\|^2 \leq d\sigma^2$$

Assumptions 1 and 2 are common in the literature [2] and hold if the data used for computing the gradients is drawn uniformly and independently.

**Assumption 3** (Linear growth of $r$-th moment).

$$\mathbb{E}_{\xi_p} \|\boldsymbol{G}(\boldsymbol{x}, \xi_p)\|^r \leq A_r + B_r \|\boldsymbol{x}\|^r \;\; \forall \boldsymbol{x} \in \mathbb{R}, \;\; r = 2, 3, 4$$

Assumption 3 translates into "the $r$-th moment of the gradient estimator grows linearly with the $r$-th power of the norm of the model" as assumed in [2].

**Assumption 4** (Lipschitz gradient).

$$||\boldsymbol{\nabla} Q(\boldsymbol{x}_1) - \boldsymbol{\nabla} Q(\boldsymbol{x}_2)|| \leq K ||\boldsymbol{x}_1 - \boldsymbol{x}_2||$$

**Assumption 5** (Convexity in the horizon). *We require that beyond a certain horizon, $\|\boldsymbol{x}\| \geq D$, there exist $\epsilon > 0$ and $0 \leq \beta < \pi/2$ such that $\|\nabla Q(\boldsymbol{x})\| \geq \epsilon > 0$ and $\frac{\langle \boldsymbol{x}, \nabla Q(\boldsymbol{x})\rangle}{\|\boldsymbol{x}\| \cdot \|\nabla Q(\boldsymbol{x})\|} \geq \cos\beta$.*

Assumption 5 is the same as in [1], which in turn is a slight refinement of a similar assumption in [2]. It essentially states that, beyond a certain horizon $D$ in the parameter space, the opposite of the gradient points towards the origin.

**Definition 2** (Byzantine resilience). *Let $Q$ be any cost function satisfying the aforementioned assumptions. Let $A$ be any distributed SGD scheme. We say that $A$ is Byzantine-resilient if the sequence $\nabla Q(\boldsymbol{x}_t) = 0$ converges almost surely to zero, despite the presence of up to $f$ Byzantine workers.*

**Theorem 1** (Optimal Slowdown). *We define the slowdown $SL$ as the ratio between the number of updates from honest workers that pass the Lipschitz filter and the total number of updates delivered at the parameter server. We derive the upper and lower bounds of $SL$ in the following.*

$$\frac{n-2f}{n-f} \leq SL \leq \frac{n-f}{n}$$

*The upper and lower bounds are tight and hold when there are $f$ Byzantine workers and no Byzantine workers respectively. Therefore Kardam achieves the optimal bounds with respect to any Byzantine-resilient SGD scheme and $n \approx 3f$ workers.*

*Proof.* Any Byzantine-resilient SGD scheme assuming $f$ Byzantine workers will at most use $\frac{n-f}{n}$ of the total available workers (upper bound). By definition, the Lipschitz filter accepts the gradients computed by $\frac{n-f}{n}$ of the total workers with empirical Lipschitzness below $\hat{K}_t$. If every worker is honest, then the filter accepts gradients from $\frac{n-f}{n}$ of the workers. We thus get the tightness of the upper bound for the slowdown of Kardam. For the lower bound, the Byzantine workers can know that putting a gradient proposition above $\hat{K}_t$ will get them filtered out and the parameter server will end up using only the honest workers available. The optimal attack would therefore be to slowdown the server by getting tiny-Lipschitz gradients accepted while preventing the model from actually changing. This way, the Byzantine workers will make the server filter gradients from a total of $f$ out of the $n-f$ honest workers, leaving only $n-2f$ useful workers for the server. □

**Theorem 2** (Byzantine resilience in asynchrony). *Let $A$ be any distributed SGD scheme. If the maximum successive gradients that $A$ accepts from a single worker and the maximum delay are both unbounded, then $A$ cannot be Byzantine-resilient when $f \geq 1$.*

*Proof.* Without any restrictions, the parameter server could only accept successive gradients from the same Byzantine worker (without getting any update from any honest worker), for example, if the Byzantine worker is faster than any other worker (which is true by the definition of a Byzantine worker and by the fact that delays on (honest) workers are unbounded). This way, the Byzantine worker can force the parameter server to follow arbitrarily bad directions and never converge. Hence, without any restriction on the number of gradients from the workers, we prove the impossibility of asynchronous Byzantine resilience. Readers familiar with distributed computing literature might note that if asynchrony was possible for Byzantine SGD without restricting the number of successive gradients from a single worker, this could be used as an abstraction to solve asynchronous Byzantine consensus (that is impossible to solve [3]). This provides another proof (by contradiction) for our theorem. □

**Theorem 3** (Correct cone and bounded statistical moments)**.** *If $N > 3f + 1$ then for any $t \geq t_r$ (we show that $t_r \in \mathcal{O}(\frac{1}{K\sqrt{|\xi|}})$ where $|\xi|$ is the batch-size of honest workers):*

$$\mathbb{E}(\|\boldsymbol{Kar}_t\|^r) \leq A'_r + B'_r\|\boldsymbol{x}_t\|^r$$

*for any $r = 2, 3, 4$ and*

$$\langle \mathbb{E}(\boldsymbol{Kar}_t), \boldsymbol{\nabla} Q_t \rangle = \Omega(1 - \frac{\sqrt{d}\sigma}{\|\boldsymbol{\nabla} Q(\boldsymbol{x}_t)\|})\|\boldsymbol{\nabla} Q(\boldsymbol{x}_t)\|^2$$

*The expectation is on the random samples used for training.*

*Proof.* First of all, it is important to note that a Byzantine worker can lie about its Lipschitz coefficient without being able to fool the parameter server. The median Lipschitz coefficient is always bounded between the Lipschitz coefficients of two correct worker, and it is against that the gradient of the Byzantine worker would be tested to be filtered out if harmful and accepted if useful.

**Lemma 1** (Limit of successive gradients)**.** *The frequency filter ensures that any sequence of length $2f + 1$ consequently accepted gradients contains at least $f + 1$ gradients computed by honest workers.*

*Proof.* Given any sequence of $2f + 1$ consequently accepted gradients ($L$), we denote by $S$ the set of workers that computed these gradients. The frequency filter guarantees that any $f$ workers in $S$ computed at most $f$ gradients in $L$. At most $f$ workers in $S$ can be Byzantine, thus at least $f + 1$ gradients in $L$ are from honest workers. □

We start the proof of Theorem 3 by proving that Kardam acts as self-stabilizing mechanism that guarantees the global confinement of the parameter vector using the following remark.

**Lemma 2** (Global Confinement)**.** *Let $\boldsymbol{x}_t$ the sequence of parameter models visited by $\boldsymbol{Kar}$. There exist a constant $D > 0$ such that the sequence $\boldsymbol{x}_t$ almost surely verifies $\|\boldsymbol{x}_t\| \leq D$ when $t \mapsto \infty$.*

*Proof.* (Global Confinement) Lemma 2 can be proven by using Remark 1 and the proof of confinement in [2].

**Remark 1.** *Let $r = 2, 3, 4$. There exist $A'_r \geq 0$ and $B'_r \geq 0$ such that:*
$(\forall t \geq 0)\mathbb{E}\|\boldsymbol{Kar}_t(\boldsymbol{x}_t, \xi)\|^r \leq A'_r + B'_r\|\boldsymbol{x}_t\|^r$

*Proof.* (Remark 1). Note that if $\boldsymbol{Kar}_t(\boldsymbol{x}_t)$ comes from a honest worker, we have $\boldsymbol{Kar}_t(\boldsymbol{x}_t, \xi) = \boldsymbol{G}(\boldsymbol{x}_t, \xi)$ therefore, $(\forall t \geq 0)\mathbb{E}\|\boldsymbol{Kar}_t(\boldsymbol{x}_t, \xi)\|^r \leq A_r + B_r\|\boldsymbol{x}_t\|^r$ since by assumption on the estimator $\boldsymbol{G}$ used by honest workers, we have $(\forall \boldsymbol{x} \in \mathbb{R})\mathbb{E}\|\boldsymbol{G}(\boldsymbol{x}, \xi)\|^r \leq A_r + B_r\|\boldsymbol{x}\|^r$.

Let $t > 2f+1$ be any epoch at the parameter server. Because of the Lipschitz filter (passed by $\boldsymbol{Kar}_t$), there exists $i \leq f$ such that $\boldsymbol{Kar}_{t-i}(\boldsymbol{x}_{t-i})$ comes from an honest worker. Therefore, $\|\boldsymbol{x}_{t-i}\| \leq \|\boldsymbol{x}_t\| + \sum_{l=1}^{i} \gamma'_{t-l}\boldsymbol{Kar}_{t-l}(\boldsymbol{x}_{t-l}) \leq \|\boldsymbol{x}_t\| + \sum_{l=1}^{i} \gamma_{t-l} \cdot \frac{\min(\boldsymbol{Kar}_0, \|\boldsymbol{Kar}_{t-l}(\boldsymbol{x}_{t-l})\|)}{\|\boldsymbol{Kar}_{t-l}(\boldsymbol{x}_{t-l})\|} \cdot \boldsymbol{Kar}_{t-l}(\boldsymbol{x}_{t-l}) \leq f \cdot \boldsymbol{Kar}_0 + \|\boldsymbol{x}_t\|$.

So, for $r = 2, 3, 4$ there exists $C_r$ such that $\|\boldsymbol{x}_{t-i}\|^r \leq (f \cdot \boldsymbol{Kar}_0)^r + C_r\|\boldsymbol{x}_t\|^r$.

According to the Lipschitz criteria:

$$
\begin{aligned}
&\|\boldsymbol{Kar}_t(\boldsymbol{x}_t)\| \\
&\leq K_t(\|\boldsymbol{x}_t\| + \|\boldsymbol{x}_{t-1}\|) + \|\boldsymbol{Kar}_{t-1}(\boldsymbol{x}_{t-1})\| \\
&\leq \sum_{l=1}^{i} K_{t-l+1}(\|\boldsymbol{x}_{t-l+1}\| + \|\boldsymbol{x}_{t-l}\|) + \|\boldsymbol{Kar}_{t-i}(\boldsymbol{x}_{t-i})\| \\
&\leq 2K\sum_{l=0}^{i} \|\boldsymbol{x}_{t-l}\| + \|\boldsymbol{Kar}_{t-i}(\boldsymbol{x}_{t-i})\| \\
&\leq 2K\sum_{l=0}^{i}\sum_{s=l}^{i-1} [\gamma'_{t-s} \cdot \|\boldsymbol{Kar}_{t-s}(\boldsymbol{x}_{t-s})\| + \|\boldsymbol{x}_{t-i}\|] + \|\boldsymbol{Kar}_{t-i}(\boldsymbol{x}_{t-i})\| \\
&\leq 2K\sum_{l=0}^{i}\sum_{s=l}^{i-1} \gamma_{t-s}\|\boldsymbol{Kar}_{t-s}(\boldsymbol{x}_{t-s})\| \cdot \frac{\min(\boldsymbol{Kar}_0, \|\boldsymbol{Kar}_{t-s}(\boldsymbol{x}_{t-s})\|)}{\|\boldsymbol{Kar}_{t-s}(\boldsymbol{x}_{t-s})\|} \\
&\quad + 2fK\|\boldsymbol{x}_{t-i}\| + \|\boldsymbol{Kar}_{t-i}(\boldsymbol{x}_{t-i})\| \\
&\leq Kf(f-1)\boldsymbol{Kar}_0 + 2fK\|\boldsymbol{x}_{t-i}\| + \|\boldsymbol{Kar}_{t-i}(\boldsymbol{x}_{t-i})\| \\
&= D + E\|\boldsymbol{x}_{t-i}\| + F\|\boldsymbol{Kar}_{t-i}(\boldsymbol{x}_{t-i})\|
\end{aligned}
$$

Where $K$ is the global Lipschitz. (We do not need to know the value of $K$ to implement $\boldsymbol{Kar}$ but we use it for the proofs.) Taking both side of the above inequality to the power $r$, we have the following for $r = 2 \dots 4$ for constants $D_r$, $E_r$ and $F_r$:

$\|\boldsymbol{Kar}_t(\boldsymbol{x}_t)\|^r \leq D_r + E_r \cdot \|\boldsymbol{x}_{t-i}\|^r + F_r \cdot E\|\boldsymbol{Kar}_{t-i}(\boldsymbol{x}_{t-i})\|^r$

As $\boldsymbol{Kar}_{t-i}(\boldsymbol{x}_{t-i})$ comes from an honest worker, using the Jensen inequality and the assumption on honest workers. We can take the expected value on $\xi$.

$$
\begin{aligned}
&\mathbb{E}\|\boldsymbol{Kar}_t(\boldsymbol{x}_t)\|^r \\
&\leq D_r + E_r \cdot \|\boldsymbol{x}_{t-i}\|^r + F_r[A_r + B_r\|\boldsymbol{x}_{t-i}\|^r] \\
&= D_r + F_rA_r + \|\boldsymbol{x}_{t-i}\|^r[E_r + F_rB_r] \\
&\leq D_r + F_rA_r + [(f \cdot \boldsymbol{Kar}_0)^r + C_r\|\boldsymbol{x}_t\|^r] \cdot [E_r + F_rB_r] \\
&= D_r + F_rA_r + f^r\boldsymbol{Kar}_0^r[E_r + F_rB_r] + [E_r + F_rB_r] \cdot \|\boldsymbol{x}_t\|^r
\end{aligned}
$$

We denote by $A'_r = D_r + F_r A_r + (f \cdot \boldsymbol{Kar}_0)^r \cdot [E_r + F_r B_r]$ and $B'_r = E_r + F_r B_r$, we obtain:
$\mathbb{E}\|\boldsymbol{Kar}_t(\boldsymbol{x}_t)\|^r \leq A'_r + B'_r \|\boldsymbol{x}_t\|^r$ □

Remark 1 shows that with $\boldsymbol{Kar}$, all the assumptions of Bottou [2] (Section 5.2) are holding even in the presence of Byzantine workers, and thus, the global confinement of $\boldsymbol{x}_t$ stated in Lemma 2. □

Remark 1 have proved the first part of Theorem 3 To continue the proof of this Theorem, the goal is to find a lower bound on the scalar product between Kardam and the real gradient of the cost. This is achieved via an upper bound on: $\|\mathbb{E}\boldsymbol{Kar}_t - \boldsymbol{\nabla} Q_t(\boldsymbol{x}_t)\|$. Let $p$ the worker whose gradient estimation $\boldsymbol{g_p}$ was selected by Kardam to be the update for epoch $t$ at the parameter server. According to Lemma 1, considering the latest $2f+1$ timestamps, at least $f+1$ of updates came from honest workers. Hence, there exists $i < f$ such that, $\boldsymbol{Kar}_{t-i}$ came from an honest worker. Hence, $\mathbb{E}\boldsymbol{Kar}_{t-i} = \boldsymbol{\nabla} Q_{t-i}$. By applying the triangle inequality twice, we have:

$$\begin{aligned} \|\boldsymbol{Kar}_t - \boldsymbol{\nabla} Q(\boldsymbol{x}_t)\| \leq & \|\boldsymbol{Kar}_t - \boldsymbol{Kar}_{t-i}\| \\ & + \|\boldsymbol{Kar}_{t-i} - \boldsymbol{\nabla} Q(\boldsymbol{x}_{t-i})\| \\ & + \|\boldsymbol{\nabla} Q(\boldsymbol{x}_{t-i}) - \boldsymbol{\nabla} Q(\boldsymbol{x}_t)\| \end{aligned}$$

We know:

$$\begin{aligned} \|\boldsymbol{Kar}_{t-i} - \boldsymbol{Kar}_t\| \leq \sum_{k=i}^{1} \|\boldsymbol{Kar}_{t-k} - \boldsymbol{Kar}_{t-k+1}\| \leq K \sum_{k=1}^{i} \|\boldsymbol{x}_{t-k+1} - \boldsymbol{x}_{t-k}\| \\ \leq K \sum_{k=1}^{i} \gamma_{t-k} \|\boldsymbol{Kar}_{t-k}\| \leq i \cdot K \cdot \gamma_{t-i} \cdot \|\boldsymbol{Kar}\|_{max(t,i)} \end{aligned}$$

where, $\|\boldsymbol{Kar}\|_{max(t,i)}$ is the upper-bound on the norm of $\boldsymbol{Kar}$ in the list from $t-i$ to $t-1$. Since $i < f$, we have $\|\boldsymbol{Kar}_{t-i} - \boldsymbol{Kar}_t\| \leq fK\gamma_{t-i}\|\boldsymbol{Kar}\|_{max(t,i)}$. Since $\boldsymbol{x}_t$ is globally confined (Lemma 2), by continuous differentiability of $Q$, so will be $\|\boldsymbol{\nabla} Q(\boldsymbol{x}_{t,i})\|$, therefore $fK\|\boldsymbol{Kar}\|_{max(t,i)}$ is bounded, and multiplies $\gamma_{t-i}$ in the right hand side of the last inequality, and we know from the hypothesis on the learning rate that $\lim_{t\to\infty} \gamma_t = 0$ (sequence of summable squares, therefore goes to zero). Since $i < f$ (and obviously, $f$, as a global variable, is independent of $t$), then we also have $\lim_{t\to\infty} \gamma_{t-i} = 0$. This means that for every $\epsilon > 0$, eventually, the left hand-side of the above inequality is bounded by $\epsilon\|\boldsymbol{Kar}_{t-i} - \boldsymbol{\nabla} Q(\boldsymbol{x}_{t-i})\|$, more precisely, since $\gamma_t$ is typically $\mathcal{O}(\frac{1}{t})$, this will hold after $t_r$ such that $t_r = \Omega(\frac{1}{\epsilon K})$.

By replacing in Formula 1, we get:

$$\begin{aligned}
\|\boldsymbol{Kar}_t - \nabla Q(\boldsymbol{x}_t)\| &\leq (1+\epsilon)\|\boldsymbol{Kar}_{t-i} - \nabla Q(\boldsymbol{x}_{t-i})\| + \|\nabla Q(\boldsymbol{x}_{t-i}) - \nabla Q(\boldsymbol{x}_t)\| \\
&\leq (1+\epsilon)\|\boldsymbol{Kar}_{t-i} - \nabla Q(\boldsymbol{x}_{t-i})\| + \sum_{s=1}^{i} K_{t-s}\gamma_{t-s}\|\nabla Q(\boldsymbol{x}_{t-s})\| \\
&\leq (1+\epsilon)\|\boldsymbol{Kar}_{t-i} - \nabla Q(\boldsymbol{x}_{t-i})\| + f \cdot K \cdot \gamma_{t-i} \cdot \|\nabla Q\|_{max(t,i)}.
\end{aligned}$$

Where $K_{t-s}$ is the real local Lipschitz coefficient of the loss function at epoch $t-s$. Let $j = \min(\frac{\sqrt{d}\sigma}{2}, \|\nabla Q(\boldsymbol{x}_t)\| - \sqrt{d}\sigma)$, $C = \frac{j}{2\epsilon\sqrt{d}\sigma}$, $\epsilon' = \frac{j}{2.C}$. As $\lim_{t\to\infty}\gamma_t = 0$ and $\|\nabla Q\|_{max(t,i)}$ is bounded, there exist a time after which, the above quantity can be made bounded as

$$\|\boldsymbol{Kar}_t - \nabla Q(\boldsymbol{x}_t)\| \leq (1+\epsilon)\|\boldsymbol{Kar}_{t-i} - \nabla Q(\boldsymbol{x}_{t-i})\| + \epsilon'.$$

And hence:

$$\begin{aligned}
\|\mathbb{E}(\boldsymbol{Kar}_t) - \nabla Q(\boldsymbol{x}_t)\| &\leq \mathbb{E}(\|\boldsymbol{Kar}_t - \nabla Q(\boldsymbol{x}_t)\|) \\
&\leq (1+\epsilon)\mathbb{E}(\|\boldsymbol{Kar}_{t-i} - \nabla Q(\boldsymbol{x}_{t-i})\|) + \epsilon'.
\end{aligned}$$

since $\boldsymbol{Kar}_{t-i}$ comes from a correct worker, we have:

$$\mathbb{E}(\|\boldsymbol{Kar}_{t-i} - \nabla Q(\boldsymbol{x}_{t-i})\|) \leq \sqrt{d}\sigma$$

Therefore, $\|\mathbb{E}(\boldsymbol{Kar}_t) - \nabla Q(\boldsymbol{x}_t)\| \leq (1+\epsilon)\sqrt{d}\sigma + \epsilon'$. Consequently, Kardam only selects vectors that live on average in the cone of radius $\alpha$ around the true gradient, where $\alpha$ is given by:

$\sin(\alpha) = \frac{(1+\epsilon)\sqrt{d}\sigma + \epsilon'}{\|\nabla Q(\boldsymbol{x}_t)\|}$. (as long as $\|\nabla Q(\boldsymbol{x}_t)\| > (1+\epsilon)\sqrt{d}\sigma + \epsilon'$, this has a sense)

Note:

- The $\sqrt{d}$ in $\|\nabla Q(\boldsymbol{x}_t)\| > \sqrt{d}\sigma$ is not a harsh requirement, we are using the conventional notation where $\sqrt{d}\sigma$ is the upper bound on the variance, $\sigma$ should be seen as the "component-wise" standard deviation, therefore, the norm of a non-trivial gradient is naturally larger than the vector-wise standard deviation of its estimator, which is typically $\sqrt{d}\sigma$.
- As long as the true gradient has *a nontrivial meaning* (it is larger than the standard deviation of its correct estimators), $\alpha$ is strictly bounded between $-\frac{\pi}{2}$ and $\frac{\pi}{2}$, which means that as long as there is no convergence to null gradients, Kardam is selecting vectors in the correct cone around the true gradient. Most importantly, this angle shrinks to zero when the variance is too small compared to the norm of the gradient, i.e., with large batch-sizes, Kardam boils down to be an unbiased gradient estimator. However, we only require the "component-wise" condition.

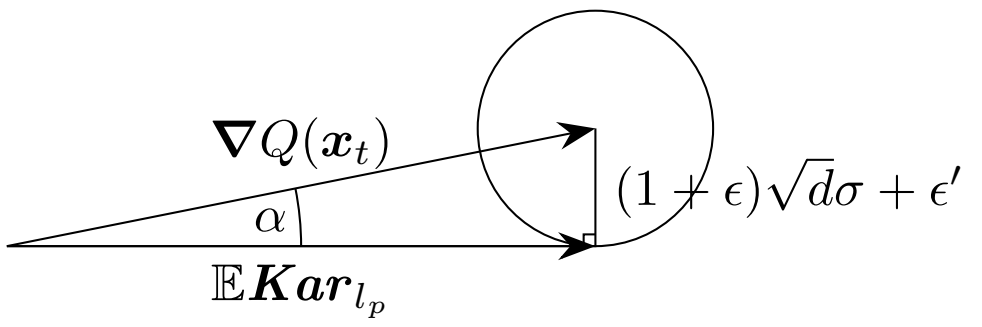

**Figure 1:** If $\left\|\mathbb{E}\boldsymbol{Kar}_{l_p} - \nabla Q(\boldsymbol{x}_t)\right\| \leq (1+\epsilon)\sqrt{d}\sigma + \epsilon'$ then $\langle \mathbb{E}\boldsymbol{Kar}_{l_p}, \nabla Q(\boldsymbol{x}_t)\rangle$ is upper bounded by $(1-\sin\alpha)\|\nabla Q(\boldsymbol{x}_t)\|^2$ where $\sin\alpha = \frac{(1+\epsilon)\sqrt{d}\sigma+\epsilon'}{\|\nabla Q(\boldsymbol{x}_t)\|}$.

In fact, as long as $\|\nabla Q(\boldsymbol{x}_t)\| > \sqrt{d}.\sigma$, we can consider small enough $\epsilon$ and $\epsilon'$ such that $D_1 = (1+\frac{3}{4C})\frac{\sqrt{d}\sigma}{\|\nabla Q(\boldsymbol{x}_t)\|}$, $D_2 = \frac{1}{C} + \frac{C-1}{C}\frac{\sqrt{d}\sigma}{\|\nabla Q(\boldsymbol{x}_t)\|}$ , and $\sin(\alpha) = \min(D_1, D_2) < 1$. This indeed guarantees that $\alpha < \frac{\pi}{2}$, moreover, it is enough to take $C >> \frac{\nabla Q(\boldsymbol{x}_t)\|}{\sqrt{d}\sigma}$ and $\alpha$ would satisfy $\sin(\alpha) \approx \frac{\sqrt{d}\sigma}{\|\nabla Q(\boldsymbol{x}_t)\|}$.

Actually, in a list of $L$ previous selected vectors, more than half of the vectors are from correct workers. (progress is made: liveness)

Consider a sublist of $L$ from $L_i$ to $L_j$. At the time of adding a worker in $L_j$, the frequency criteria was checked for the new addition to $L$. The active table at that time assure that in any new sublist of $L$, especially $L_i^j$), any $f$ workers appear at most $\frac{j-i}{2}$ times. As the number of Byzantine workers is maximum $f$. in sublist $L_i^j$, the Byzantine workers did less than half of the updates. In other words, at least half of the updates come from honest workers. This proves the safety of Kardam.

The Byzantine workers may stop sending updates or send incorrect updates. In the case where the Byzantine workers stop sending updates, Kardam still guarantees liveness. The reason is that there are at least $2f+1$ honest workers who update the model. □

# 2 Convergence Analysis

**Definition 3** (Dampening function). *We employ a bijective and strictly decreasing dampening function $\tau \mapsto \Lambda(\tau)$ with $\Lambda(0) = 1$.*[1] *Note that every bijective function is also invertible, i.e., $\Lambda^{-1}(\nu)$ exists for every $\nu$ in the range of the $\Lambda$ function.*

Let $\boldsymbol{\Lambda}_t$ be the set of $\Lambda$ values associated with the gradients at epoch $t$.

$$\boldsymbol{\Lambda}_t = \{\Lambda(\tau_{tl}) \mid [g,l] \in \mathcal{G}_t\}$$

We partition the set $\mathcal{G}_t$ of gradients at epoch $t$ according to their $\Lambda$-value as follows.

$$\mathcal{G}_t = \bigsqcup_{\lambda \in \boldsymbol{\Lambda}_t} \mathcal{G}_{t\lambda}$$

$$\mathcal{G}_{t\lambda} = \{[g,l] \in \mathcal{G}_t \mid \Lambda(\tau_{tl}) = \lambda\}$$

[1] If $\Lambda(0) = 1$, then there is no decay for gradients computed on the latest version of the model, i.e., $\tau_{tl} = 0$.

Therefore, the update equation can be reformulated as follows.

$$\boldsymbol{x}_{t+1} = \boldsymbol{x}_t - \gamma_t \sum_{\lambda \in \boldsymbol{\Lambda}_t} \lambda \cdot \sum_{[\boldsymbol{G}(\boldsymbol{x}_l;\xi),l] \in \mathcal{G}_{t\lambda}} \boldsymbol{G}(\boldsymbol{x}_l;\xi)$$

**Definition 4** (Adaptive learning rate)**.** *Given the Lipschitz constant $K$, the total number of epochs $T$, and the total number of gradients in each epoch as $M$, we define $\gamma_t$ as follows.*

$$\gamma_t = \underbrace{\sqrt{\frac{Q(\boldsymbol{x}_1) - Q(\boldsymbol{x}^*)}{K \cdot T \cdot M \cdot d \cdot \sigma^2}}}_{\gamma} \cdot \underbrace{\frac{M}{\sum_{\lambda \in \boldsymbol{\Lambda}_t} \lambda \cdot |\mathcal{G}_{t\lambda}|}}_{\mu_t} \tag{1}$$

where $\gamma$ is the baseline component of the learning rate and $\mu_t$ is the adaptive component that depends on the amount of stale updates that the server receives at epoch $t$. Moreover, $\mu_t$ incorporates the total staleness at any epoch $t$ based on the staleness coefficients ($\lambda$) associated with all the gradients received in epoch $t$.

**Comments on $Q(\boldsymbol{x}^*)$.** $\boldsymbol{x}^*$ refers to the (not necessarily global) optimum we are heading to, and on which our adaptive learning rate depends. Assuming this value is known was made just for the sake of a proof, as is usually done in proofs for the speed of convergence of SGD (e.g., the references provided by the reviewer). In practice, one does not need to know $Q(\boldsymbol{x}^*)$ and can assume it to be lower bounded (Bottou1998). This will produce overshooting (large steps) in the early phases of Kardam, but will get to small enough step sizes: The baseline part of our adaptive learning rate contains a term $1/T$, where $T$ is the total number of iterations (also unknown before we run SGD). In practice, it is replaced by $1/t$ ($t$: epoch at the server). This part of our learning rate decreases with $t$ and will compensate for the overshooting described above (overcoming the overshooting in at most $O(1/Q(x_1))$ steps).

**Remark 2** (Correct cone)**.** *As a consequence of passing the filter and of Theorem 3, $\boldsymbol{G}$ satisfies the following.*

$$\langle \mathbb{E}_\xi \boldsymbol{G}(\boldsymbol{x};\xi), \boldsymbol{\nabla} Q(\boldsymbol{x}) \rangle > \Omega\left( (\|\boldsymbol{\nabla} Q(x_t)\| - \sqrt{d}\sigma) \|\boldsymbol{\nabla} Q(x_t)\| \right)$$

The theoretical guarantee for the convergence rate of Kardam depends on Assumptions 2,4 and Remark 2. These assumptions are weaker than the convergence guarantees in [7, 4]. In particular, due to unbounded delays and the potential presence of Byzantine workers, we only assume the unbiased gradient estimator $\boldsymbol{G}(\cdot)$ for honest workers (Assumption 1). We instead employ (Remark 2) the fact that $\boldsymbol{G}(\cdot)$ and $\boldsymbol{\nabla} Q(\boldsymbol{x})$ make a lower bounded angle together (and subsequently a lower bounded scalar product) for all the workers. The classical unbiased assumption is more restrictive as it requires this angle to be exactly equal to 0, and the scalar product to be equal to $\|\boldsymbol{\nabla} Q(\boldsymbol{x})\| \cdot \|\boldsymbol{G}(\boldsymbol{x})\|$. Most importantly, we highlight the fact that those assumption are satisfied by Kardam, since every gradient used in this section to compute the ***Kar*** update has passed the Lipschitz filter of the previous section.

**Theorem 4** (Convergence guarantee). *We provide the convergence guarantee in terms of the ergodic convergence that is the weighted average of the $\mathcal{L}_2$ norm of all gradients ($||\boldsymbol{\nabla} Q(\boldsymbol{x}_t)||^2$). Using the above-mentioned assumptions, and the maximum adaptive rate $\mu_{\max} = \max\{\mu_1, \ldots, \mu_t\}$, we have the following bound on the ergodic convergence rate.*

$$\frac{1}{T}\sum_{t=1}^{T}\mathbb{E}\|\boldsymbol{\nabla} Q(\boldsymbol{x}_t)\|^2 \leq (2+\mu_{\max}+\gamma K M \chi \mu_{\max}) \cdot \gamma K \cdot d\sigma^2 + d \cdot \sigma^2 + 2DK\sigma\sqrt{d} + K^2 D^2 \tag{2}$$

*under the prerequisite that*

$$\sum_{\lambda \in \boldsymbol{\Lambda}_t} \left\{ K\gamma_t^2 |\boldsymbol{\Lambda}_t| + \sum_{s=1}^{\infty} \sum_{\nu \in \boldsymbol{\Lambda}_{t+s}} \gamma_{t+s} K^2 \nu |\mathcal{G}_{t+s,\nu}| \Lambda^{-1}(\nu) \mathbb{I}_{(s \leq \Lambda^{-1}(\nu))} \gamma_t^2 |\boldsymbol{\Lambda}_t| \right\} \lambda^2 \leq \sum_{\lambda \in \boldsymbol{\Lambda}_t} \frac{\gamma_t \lambda}{|\mathcal{G}_{t\lambda}|} \tag{3}$$

*where the Iverson indicator function is defined as follows.*

$$\mathbb{I}_{(s \leq \Delta)} = \begin{cases} 1 & \text{if } s \leq \Delta \\ 0 & \text{otherwise.} \end{cases}$$

It is important to note that the prerequisite (Inequality 3) holds for any decay function $\Lambda$ (since $\lambda < 1$ holds by definition) and for any standard learning rate schedule such that $\gamma_t < 1$. Various GD approaches [7, 5, 6, 4] provide convergence guarantees with similar prerequisites.

*Proof.* We provide the convergence guarantee in terms of *ergodic convergence*—the weighted average of the $\mathcal{L}_2$ norm of all gradients ($||\boldsymbol{\nabla} Q(\boldsymbol{x}_t)||^2$). For the sake of clarity in the proofs, if $X$ is a set, we also denote its cardinality by $X$.

**Lemma 3.** *1 Assume that, for all epochs $1 \leq t \leq T$*

$$\sum_{\lambda \in \boldsymbol{\Lambda}_t} \left\{ K\gamma_t^2 |\boldsymbol{\Lambda}_t| + \sum_{s=1}^{\infty} \sum_{\nu \in \boldsymbol{\Lambda}_{t+s}} \gamma_{t+s} K^2 \nu |\mathcal{G}_{t+s,\nu}| \Lambda^{-1}(\nu) \mathbb{I}_{(s \leq \Lambda^{-1}(\nu))} \gamma_t^2 |\boldsymbol{\Lambda}_t| \right\} \lambda^2 \leq \sum_{\lambda \in \boldsymbol{\Lambda}_t} \frac{\gamma_t \lambda}{|\mathcal{G}_{t\lambda}|}$$

*Then, the ergodic convergence rate is bounded as follows.*

$$\frac{\sum_{t=1}^{T}\left(\gamma_t \sum_{\lambda\in\boldsymbol{\Lambda}_t} \lambda \mathcal{G}_{t\lambda}\right)\mathbb{E}||\boldsymbol{\nabla} Q(\boldsymbol{x}_t)||^2}{\sum_{t=1}^{T}\gamma_t \sum_{\lambda\in\boldsymbol{\Lambda}_t} \lambda \mathcal{G}_{t\lambda}} \leq \frac{2(Q(\boldsymbol{x}_1)-Q(\boldsymbol{x}^*))}{\sum_{t=1}^{T}\gamma_t \sum_{\lambda\in\boldsymbol{\Lambda}_t} \lambda \mathcal{G}_{t\lambda}}$$
$$+\frac{\left(\sum_{t=1}^{T} K\gamma_t^2 \sum_{\lambda\in\boldsymbol{\Lambda}_t}\lambda^2\mathcal{G}_{t\lambda} + \gamma_t K^2 \sum_{\lambda\in\boldsymbol{\Lambda}_t}\lambda\mathcal{G}_{t\lambda} \sum_{j=t-\Lambda^{-1}(\lambda)}^{t-1}\gamma_j^2 \sum_{\lambda'\in\boldsymbol{\Lambda}_j}\lambda'^2\mathcal{G}_{j\lambda'}\right)\cdot d\cdot\sigma^2}{\sum_{t=1}^{T}\gamma_t \sum_{\lambda\in\boldsymbol{\Lambda}_t} \lambda \mathcal{G}_{t\lambda}}$$

**Remark 3.** *Given a list of vectors $u_1, \ldots, u_N$, we implicitly use the following inequality in our proof.*

$$\left\|\sum_{i=1}^{N} u_i\right\|^2 \leq N\cdot\sum_{i=1}^{N}\|u_i\|^2 \tag{3}$$

*Proof.* For the sake of concision, for every $m=[g,l]\in\mathcal{G}_{t\lambda}$, we denote by $\xi_{[t]}$ the set of $\xi$ values that the server sends during epoch $t$. Let $\xi_{[t,*\neq m]}$ denote the set $\xi_{[t]}$ minus the variable $\xi$ corresponding to message $m$. Additionally, $G[tm] \triangleq \boldsymbol{G}(\boldsymbol{x}_{t-\tau_{tl}};\xi)$ and $\boldsymbol{\nabla} Q[tm] \triangleq \boldsymbol{\nabla} Q(\boldsymbol{x}_{t-\tau_{tl}})$.

A second order expansion of $Q$, followed by the application of the Lipschitz inequality to $\boldsymbol{\nabla} Q$ yields the following.

$$Q(\boldsymbol{x}_{t+1}) - Q(\boldsymbol{x}_t) \leq \langle \boldsymbol{\nabla} Q(\boldsymbol{x}_t), \boldsymbol{x}_{t+1}-\boldsymbol{x}_t\rangle + \frac{K}{2}\|\boldsymbol{x}_{t+1}-\boldsymbol{x}_t\|^2$$
$$\leq -\gamma_t\sum_{\boldsymbol{\Lambda}_t}\lambda\mathcal{G}_{t\lambda}\langle\boldsymbol{\nabla} Q(\boldsymbol{x}_t), \frac{1}{\mathcal{G}_{t\lambda}}\sum_{\mathcal{G}_{t\lambda}} G[tm]\rangle + \frac{K}{2}\gamma_t^2\left\|\sum_{\boldsymbol{\Lambda}_t}\lambda\sum_{\mathcal{G}_{t\lambda}} G[tm]\right\|^2$$

Taking the expectation and using the correct cone property, we have:

$$\mathbb{E}_{\xi_{[t]}} Q(\boldsymbol{x}^{(t+1)}) - Q(\boldsymbol{x}_t) \leq -\gamma_t\sum_{\boldsymbol{\Lambda}_t}\lambda\mathcal{G}_{t\lambda}\langle\boldsymbol{\nabla} Q(\boldsymbol{x}_t), \frac{1}{\mathcal{G}_{t\lambda}}\sum_{\mathcal{G}_{t\lambda}}\boldsymbol{\nabla} Q[tm]\rangle$$
$$+\frac{K}{2}\gamma_t^2\mathbb{E}_{\xi_{[t]}}\left\|\sum_{\boldsymbol{\Lambda}_t}\lambda\sum_{\mathcal{G}_{t\lambda}} G[tm]\right\|^2$$

Using $\langle a, b\rangle = \frac{||a||^2+||b||^2-||a-b||^2}{2}$, we obtain the following inequality.

$$\mathbb{E}_{\xi_{[t]}} Q(\boldsymbol{x}_{t+1}) - Q(\boldsymbol{x}_t) \leq -\frac{\gamma_t}{2}\sum_{\boldsymbol{\Lambda}_t}\lambda\mathcal{G}_{t\lambda}\left\|\boldsymbol{\nabla} Q(\boldsymbol{x}_t)\right\|^2$$

$$-\frac{\gamma_t}{2}\sum_{\boldsymbol{\Lambda}_t}\lambda\mathcal{G}_{t\lambda}\left\|\frac{1}{\mathcal{G}_{t\lambda}}\sum_{\mathcal{G}_{t\lambda}}\boldsymbol{\nabla} Q[tm]\right\|^2 + \frac{K\gamma_t^2}{2}\underbrace{\mathbb{E}_{\xi_{[t]}}\left\|\sum_{\boldsymbol{\Lambda}_t}\lambda\sum_{\mathcal{G}_{t\lambda}} G[tm]\right\|^2}_{S_1}$$

$$+\frac{\gamma_t}{2}\sum_{\boldsymbol{\Lambda}_t}\lambda\mathcal{G}_{t\lambda}\underbrace{\left\|\boldsymbol{\nabla} Q(\boldsymbol{x}_t) - \frac{1}{\mathcal{G}_{t\lambda}}\sum_{\mathcal{G}_{t\lambda}}\boldsymbol{\nabla} Q[tm]\right\|^2}_{S_2}$$

We now define two terms $S_1$ and $S_2$ as follows.

$$\begin{aligned}
S_1 &= \mathbb{E}_{\xi_{[t]}}\left\|\sum_{\boldsymbol{\Lambda}_t}\lambda\sum_{\mathcal{G}_{t\lambda}}(G[tm]-\boldsymbol{\nabla} Q[tm]) + \sum_{\boldsymbol{\Lambda}_t}\lambda\sum_{\mathcal{G}_{t\lambda}}\boldsymbol{\nabla} Q[tm]\right\|^2 \\
&= \mathbb{E}_{\xi_{[t]}}\left\|\sum_{\boldsymbol{\Lambda}_t}\lambda\sum_{\mathcal{G}_{t\lambda}}(G[tm]-\boldsymbol{\nabla} Q[tm])\right\|^2 + \mathbb{E}_{\xi_{[m]}}\left\|\sum_{\boldsymbol{\Lambda}_t}\lambda\sum_{\mathcal{G}_{t\lambda}}\boldsymbol{\nabla} Q[tm]\right\|^2 \\
&+ 2\mathbb{E}_{\xi_{[t]}}\langle\sum_{\boldsymbol{\Lambda}_t}\lambda\sum_{\mathcal{G}_{t\lambda}}(G[tm]-\boldsymbol{\nabla} Q[tm]), \sum_{\boldsymbol{\Lambda}_t}\lambda\sum_{\mathcal{G}_{t\lambda}}\boldsymbol{\nabla} Q[tm]\rangle \\
&= \mathbb{E}_{\xi_{[t]}}\left\|\sum_{\boldsymbol{\Lambda}_t}\lambda\sum_{\mathcal{G}_{t\lambda}}(G[tm]-\boldsymbol{\nabla} Q[tm])\right\|^2 + \mathbb{E}_{\xi_{[t]}}\left\|\sum_{\boldsymbol{\Lambda}_t}\lambda\sum_{\mathcal{G}_{t\lambda}}\boldsymbol{\nabla} Q[tm]\right\|^2 \\
&+ 2\langle\sum_{\boldsymbol{\Lambda}_t}\lambda\sum_{\mathcal{G}_{t\lambda}}(\boldsymbol{\nabla} Q[tm]-\boldsymbol{\nabla} Q[tm]), \sum_{\boldsymbol{\Lambda}_t}\lambda\sum_{\mathcal{G}_{t\lambda}}\boldsymbol{\nabla} Q[tm]\rangle \\
&= \underbrace{\mathbb{E}_{\xi_{[t]}}\left\|\sum_{\boldsymbol{\Lambda}_t}\lambda\sum_{\mathcal{G}_{t\lambda}}(G[tm]-\boldsymbol{\nabla} Q[tm])\right\|^2}_{A_1} + \underbrace{\mathbb{E}_{\xi_{[t]}}\left\|\sum_{\boldsymbol{\Lambda}_t}\lambda\sum_{\mathcal{G}_{t\lambda}}\boldsymbol{\nabla} Q[tm]\right\|^2}_{A_2}
\end{aligned}$$

Regarding $A_2$, applying Equation 3 yields the following inequality.

$$A_2 \leq \mathbb{E}_{\xi_{[t]}}\boldsymbol{\Lambda}_t\cdot\sum_{\boldsymbol{\Lambda}_t}\lambda^2\|\sum_{\mathcal{G}_{t\lambda}}\boldsymbol{\nabla} Q[tm]\|^2 \leq \boldsymbol{\Lambda}_t\cdot\sum_{\boldsymbol{\Lambda}_t}\lambda^2\mathbb{E}_{\xi_{[t]}}\|\sum_{\mathcal{G}_{t\lambda}}\boldsymbol{\nabla} Q[tm]\|^2$$

Regarding $A_1$, the term $\|\ldots\|^2$ is expressed as a scalar product and expanded as follows.

$$
\begin{aligned}
A_1 &= \mathbb{E}_{\xi_{[t]}} \sum_{\lambda,\lambda' \in \mathbf{\Lambda}_t} \Big( \sum_{\substack{m \in \mathcal{G}_{t\lambda}, \\ m' \in \mathcal{G}_{t\lambda'}}} \lambda\lambda' \cdot \langle G[tm] - \nabla Q[tm], G[tm'] - \nabla Q[tm'] \rangle \Big) \\
&= \text{diagonal } + \text{off-diagonal} \\
&= \sum_{\lambda \in \mathbf{\Lambda}_t} \sum_{m \in \mathcal{G}_{t\lambda}} \lambda^2 \cdot \mathbb{E}_{\xi_{[t]}} \|G[tm] - \nabla Q[tm]\|^2 + \mathbb{E}_{\xi_{[t,m' \neq m]}} \left( \mathbb{E}_{\xi} \langle G[tm] - \nabla Q[tm], G[tm'] - \nabla Q[tm'] \rangle \right) \\
&\leq \sum_{\mathbf{\Lambda}_t} \lambda^2 \mathcal{G}_{t\lambda} \cdot d \cdot \sigma^2 + d \cdot \sigma^2 + 2DK\sigma\sqrt{d} + K^2 D^2
\end{aligned}
$$

The sum over the off-diagonal terms (i.e., $(\lambda, m) \neq (\lambda', m')$) is bounded by $d \cdot \sigma^2 + 2DK\sigma\sqrt{d} + K^2D^2$. Moreover, if $\lambda \neq \lambda'$, then $m \neq m'$ because $\mathcal{G}_{t\lambda}$ and $\mathcal{G}_{t\lambda'}$ are disjoint sets and thus for any off-diagonal pair $(\lambda, m), (\lambda, m')$ we have $m \neq m'$.

$$
\begin{aligned}
&\mathbb{E}_{\xi_{[t]}} \langle G[tm] - \nabla Q[tm], G[tm'] - \nabla Q[tm'] \rangle \\
&\quad = \mathbb{E}_{\xi_{[t,m' \neq m]}} \left( \mathbb{E}_{\xi} \langle G[tm] - \nabla Q[tm], G[tm'] - \nabla Q[tm'] \rangle \right) \\
&\quad = \mathbb{E}_{\xi_{[t,m' \neq m]}} \langle \mathbb{E}_{\xi} G[tm] - \nabla Q[tm], G[tm'] - \nabla Q[tm'] \rangle \\
&\quad = \mathbb{E}_{\xi_{[t,m' \neq m]}} \left( \langle \mathbb{E}_{\xi} G[tm], G[tm'] \rangle - \langle \nabla Q[tm], G[tm'] \rangle - \langle \mathbb{E}_{\xi} G[tm], \nabla Q[tm'] \rangle + \langle \nabla Q[tm], \nabla Q[tm'] \rangle \right) \\
&\quad \leq \mathbb{E}_{\xi_{[t,m' \neq m]}} \big( \|\mathbb{E}_{\xi} G[tm]\| \cdot \|G[tm']\| + \|\nabla Q[tm]\| \cdot \|G[tm']\| \\
&\qquad + \|\mathbb{E}_{\xi} G[tm]\| \cdot \|\nabla Q[tm']\| + \|\nabla Q[tm]\| \cdot \|\nabla Q[tm']\| \big) \\
&\quad \leq d \cdot \sigma^2 + 2DK\sigma\sqrt{d} + K^2 D^2
\end{aligned}
$$

Hence, we obtain the following inequalities for $S_1$ and $S_2$.

$$
S_1 \leq \sum_{\mathbf{\Lambda}_t} \lambda^2 \mathcal{G}_{t\lambda} \cdot d \cdot \sigma^2 + \mathbf{\Lambda}_t \cdot \sum_{\mathbf{\Lambda}_t} \lambda^2 \mathbb{E}_{\xi_{[t]}} \|\sum_{\mathcal{G}_{t\lambda}} \nabla Q[tm]\|^2 + d \cdot \sigma^2 + 2DK\sigma\sqrt{d} + K^2 D^2
$$

$$
S_2 \leq \left\| \frac{1}{\mathcal{G}_{t\lambda}} \sum_{\mathcal{G}_{t\lambda}} \nabla Q(\boldsymbol{x}_t) - \nabla Q[tm] \right\|^2
$$

Recall that, since $m = [g, l] \in \mathcal{G}_{t\lambda}$, we have $\nabla Q[tm] = \nabla Q(\boldsymbol{x}_{t-\tau_{tl}})$. By applying the Lipschitz inequality, we get:

$$S_2 \le K^2 \|\boldsymbol{x}_t - \boldsymbol{x}_{t-\Lambda^{-1}(\lambda)}\|^2$$

$$\le K^2 \left\| \sum_{j=t-\Lambda^{-1}(\lambda)}^{t-1} \boldsymbol{x}_{j+1} - \boldsymbol{x}_j \right\|^2 \le K^2 \left\| \sum_{j=t-\Lambda^{-1}(\lambda)}^{t-1} \gamma_j \sum_{\nu \in \boldsymbol{\Lambda}_j} \nu \sum_{\mathcal{G}_{j\nu}} G[jm] \right\|^2$$

$$\le K^2 \underbrace{\left\| \sum_{j=t-\Lambda^{-1}(\lambda)}^{t-1} \gamma_j \sum_{\nu \in \boldsymbol{\Lambda}_j} \nu \sum_{\mathcal{G}_{j\nu}} (G[jm] - \boldsymbol{\nabla} Q[jm]) \right\|^2}_{S_3 = \|a\|^2}$$

$$+ K^2 \underbrace{\left\| \sum_{j=t-\Lambda^{-1}(\lambda)}^{t-1} \gamma_j \sum_{\nu \in \boldsymbol{\Lambda}_j} \nu \sum_{\mathcal{G}_{j\nu}} \boldsymbol{\nabla} Q[jm] \right\|^2}_{S_4 = \|b\|^2} + 2K^2 \langle a, b \rangle$$

Hence, we obtain the following inequalities for $S_3$ and $S_4$.

$$\mathbb{E}_{\xi_{[j]},\ldots} S_3 \le \sum_{j=t-\Lambda^{-1}(\lambda)}^{t-1} \gamma_j^2 \sum_{\boldsymbol{\Lambda}_j} \nu^2 \mathcal{G}_{j\nu} \cdot d \cdot \sigma^2 \quad \text{(cross-products vanish)}$$

$$\mathbb{E}_{\xi_{[j]},\ldots} S_4 \le \Lambda^{-1}(\lambda) \sum_{j=k-\Lambda^{-1}(\lambda)}^{t-1} \gamma_j^2 \boldsymbol{\Lambda}_j \sum_{\boldsymbol{\Lambda}_j} \nu^2 \mathbb{E} \left\| \sum_{\mathcal{G}_{j\nu}} \boldsymbol{\nabla} Q[jm'] \right\|^2 \quad \text{(by Eq. 3).}$$

Moreover, we have $\mathbb{E}_* \langle a, b \rangle = \langle \mathbb{E}_* a, b \rangle = 0$.

$$\mathbb{E} S_2 \le K^2 \sum_{j=t-\Lambda^{-1}(\lambda)}^{t-1} \gamma_j^2 \sum_{\boldsymbol{\Lambda}_j} \nu^2 \mathcal{G}_{j\nu} \cdot d \cdot \sigma^2$$

$$+ K^2 \Lambda^{-1}(\lambda) \sum_{j=t-\Lambda^{-1}(\lambda)}^{t-1} \gamma_j^2 \boldsymbol{\Lambda}_j \sum_{\boldsymbol{\Lambda}_j} \nu^2 \mathbb{E} \left\| \sum_{\mathcal{G}_{j\nu}} \boldsymbol{\nabla} Q[jm'] \right\|^2$$

$$\mathbb{E}_{\xi_{[t]}} Q(\boldsymbol{x}_{t+1}) - Q(\boldsymbol{x}_t) \le -\frac{\gamma_t}{2} \sum_{\boldsymbol{\Lambda}_t} \lambda \mathcal{G}_{t\lambda} \|\boldsymbol{\nabla} Q(\boldsymbol{x}_t)\|^2$$

$$+ \sum_{\boldsymbol{\Lambda}_t} \left( \frac{K\gamma_t^2 \boldsymbol{\Lambda}_t \lambda^2}{2} - \frac{\gamma_t \lambda}{2\mathcal{G}_{t\lambda}} \right) \mathbb{E} \left\| \sum_{\mathcal{G}_{t\lambda}} \boldsymbol{\nabla} Q[tm] \right\|^2$$

$$+ \left( \frac{K\gamma_t^2}{2} \sum_{\boldsymbol{\Lambda}_t} \lambda^2 \mathcal{G}_{t\lambda} + \frac{\gamma_t K^2}{2} \sum_{\boldsymbol{\Lambda}_t} \lambda \mathcal{G}_{t\lambda} \sum_{j=t-\Lambda^{-1}(\lambda)}^{t-1} \gamma_j^2 \sum_{\boldsymbol{\Lambda}_j} \nu^2 \mathcal{G}_{j\nu} \right) \cdot d \cdot \sigma^2$$

$$+ \frac{\gamma_t K^2}{2} \sum_{\boldsymbol{\Lambda}_t} \lambda \mathcal{G}_{t\lambda} \Lambda^{-1}(\lambda) \sum_{j=t-\Lambda^{-1}(\lambda)}^{t-1} \gamma_j^2 \boldsymbol{\Lambda}_j \sum_{\boldsymbol{\Lambda}_j} \nu^2 \mathbb{E} \left\| \sum_{j\nu} \boldsymbol{\nabla} Q[jm'] \right\|^2$$

Summing for $t = 1, \ldots, T$, we arrive at the following inequality.

$$\mathbb{E}Q(\boldsymbol{x}_{t+1}) - Q(\boldsymbol{x}_1) \leq -\sum_t \frac{1}{2}\left(\gamma_t \sum_{\boldsymbol{\Lambda}_t} \lambda \mathcal{G}_{t\lambda}\right) \|\boldsymbol{\nabla} Q(\boldsymbol{x}_t)\|^2$$

$$+\sum_t \left(\frac{K\gamma_t^2}{2}\sum_{\boldsymbol{\Lambda}_t}\lambda^2\mathcal{G}_{t\lambda} + \frac{\gamma_t K^2}{2}\sum_{\boldsymbol{\Lambda}_t}\lambda\mathcal{G}_{t\lambda}\sum_{j=t-\Lambda^{-1}(\lambda)}^{t-1}\gamma_j^2\sum_{\boldsymbol{\Lambda}_j}\nu^2\mathcal{G}_{j\nu}\right)\cdot d\cdot\sigma^2$$

$$+\sum_t\sum_{\boldsymbol{\Lambda}_t}\left(\frac{K\gamma_t^2\boldsymbol{\Lambda}_t\lambda^2}{2} - \frac{\gamma_t\lambda}{2\mathcal{G}_{t\lambda}}\right)\mathbb{E}\left\|\sum_{\mathcal{G}_{t\lambda}}\boldsymbol{\nabla}Q[tm]\right\|^2$$

$$+\sum_t\left(\sum_{s=1}^{\infty}\sum_{\boldsymbol{\Lambda}_{t+s}}\gamma_{t+s}K^2\nu\mathcal{G}_{t+s,\nu}\Lambda^{-1}(\nu)\mathbb{I}(s\leq\Lambda^{-1}(\nu))\right)\frac{\gamma_t\boldsymbol{\Lambda}_t\lambda^2}{2}\mathbb{E}\left\|\sum_{\mathcal{G}_{t\lambda}}\boldsymbol{\nabla}Q[tm]\right\|^2$$

The last term comes from the following observation.

$$\sum_{t=1}^{T}\sum_{\boldsymbol{\Lambda}_t}\sum_{s=1}^{\infty}Q_t^\lambda Z_{t-s}\mathbb{I}(s\leq\Lambda^{-1}(\lambda)) = \sum_{s=1}^{\infty}\sum_{t=1}^{T}\sum_{\boldsymbol{\Lambda}_t}Q_t^\lambda Z_{t-s}\mathbb{I}(s\leq\Lambda^{-1}(\lambda))$$

$$=\sum_{s=1}^{\infty}\sum_{l=1-s}^{T-s}\sum_{\boldsymbol{\Lambda}_{l+s}}Q_{l+s}^\lambda Z_l\mathbb{I}(s\leq\Lambda^{-1}(\lambda)) = \sum_{s=1}^{\infty}\sum_{t=1}^{T}\sum_{\boldsymbol{\Lambda}_{t+s}}Q_{t+s}^\lambda Z_t\mathbb{I}(s\leq\Lambda^{-1}(\lambda))$$

$$=\sum_{t=1}^{T}\left(\sum_{s=1}^{\infty}\sum_{\boldsymbol{\Lambda}_{t+s}}Q_{t+s}^\lambda\mathbb{I}(s\leq\Lambda^{-1}(\lambda))\right)Z_t$$

Since the two last terms sum to a non-positive value, we arrive at the following inequality.

$$\sum_t\frac{1}{2}\left(\gamma_t\sum_{\boldsymbol{\Lambda}_t}\lambda\mathcal{G}_{t\lambda}\right)\|\boldsymbol{\nabla}Q(\boldsymbol{x}_t)\|^2 \leq Q(\boldsymbol{x}_1) - Q(\boldsymbol{x}^*)$$

$$+\sum_t\left(\frac{K\gamma_t^2}{2}\sum_{\boldsymbol{\Lambda}_t}\lambda^2\mathcal{G}_{t\lambda} + \frac{\gamma_t K^2}{2}\sum_{\boldsymbol{\Lambda}_t}\lambda\mathcal{G}_{t\lambda}\sum_{j=t-\Lambda^{-1}(\lambda)}^{t-1}\gamma_j^2\sum_{\boldsymbol{\Lambda}_j}\nu^2\mathcal{G}_{j\nu}\right)\cdot d\cdot\sigma^2 + \mathcal{O}(\frac{1}{K\cdot\sqrt{|\xi|}})$$

□

We first recall Definition 4, which introduces the adaptive learning rate schedule, before we prove Theorem 4 via employing Lemma 3. Due to the choice of the learning rate (Definition 4), the inequality in Theorem 4 reduces to the following inequality.

$$\frac{1}{T}\sum_{t=1}^{T}\mathbb{E}\|Q(\boldsymbol{x}_t)\|^2 \leq S_5 + S_6 + S_7$$

First, we obtain the following equality for $S_5$.

$$S_5 = \frac{2(Q(\boldsymbol{x}_1) - Q(\boldsymbol{x}^*))}{\sum_{t=1}^{T} \gamma_t \sum_{\lambda \in \boldsymbol{\Lambda}_t} \lambda \mathcal{G}_{t\lambda}} = \frac{2\gamma^2 KTM \cdot d \cdot \sigma^2}{\gamma TM} = 2\gamma K \cdot d \cdot \sigma^2$$

Regarding $S_6$, we obtain the following inequality.

$$S_6 = \frac{\sum_{t=1}^{T} K\gamma_t^2 \sum_{\lambda \in \boldsymbol{\Lambda}_t} \lambda^2 \mathcal{G}_{t\lambda}}{\sum_{t=1}^{T} \gamma_t \sum_{\lambda \in \boldsymbol{\Lambda}_t} \lambda \mathcal{G}_{t\lambda}} \cdot d \cdot \sigma^2 = \frac{K\gamma^2 \sum_{t=1}^{T} \mu_t^2 \sum_{\lambda \in \boldsymbol{\Lambda}_t} \lambda^2 \mathcal{G}_{t\lambda}}{\gamma TM} \cdot d \cdot \sigma^2$$

$$\leq \frac{K\gamma^2 \sum_{t=1}^{T} \mu_t^2 \sum_{\lambda \in \boldsymbol{\Lambda}_t} \lambda \mathcal{G}_{t\lambda}}{\gamma TM} \cdot d \cdot \sigma^2 \text{ (since } \lambda^2 \leq \lambda \leq 1)$$

$$\leq \frac{K\gamma^2 TM \mu_{\max}}{\gamma TM} \cdot d \cdot \sigma^2 = \mu_{\max} \gamma K \cdot d \cdot \sigma^2$$

Finally, we obtain the following inequality for $S_7$.

$$S_7 = \frac{\sum_{t=1}^{T} \gamma_t K^2 \sum_{\lambda \in \boldsymbol{\Lambda}_t} \lambda \mathcal{G}_{t\lambda} \sum_{j=t-\Lambda^{-1}(\lambda)}^{t-1} \gamma_j^2 \sum_{\lambda' \in \boldsymbol{\Lambda}_j} \lambda'^2 M_{j\lambda'}}{\gamma TM} \cdot d \cdot \sigma^2$$

$$\leq \frac{K^2 \gamma^3 \sum_{t=1}^{T} \mu_t \sum_{\lambda \in \boldsymbol{\Lambda}_t} \lambda \mathcal{G}_{t\lambda} \sum_{j=t-\Lambda^{-1}(\lambda)}^{t-1} \mu_j^2 \sum_{\lambda' \in \boldsymbol{\Lambda}_j} \lambda'^2 M_{j\lambda'}}{\gamma TM} \cdot d \cdot \sigma^2$$

$$\leq \frac{K^2 \gamma^3 \sum_{t=1}^{T} \sum_{\lambda \in \boldsymbol{\Lambda}_t} \lambda \mathcal{G}_{t\lambda} M \Lambda^{-1}(\lambda) \mu_{\max}}{\gamma TM} \cdot d \cdot \sigma^2$$

$$\leq \frac{K^2 \gamma^3 \sum_{t=1}^{T} \sum_{\lambda \in \boldsymbol{\Lambda}_t} \mathcal{G}_{t\lambda} M \chi \mu_{\max}}{\gamma TM} \cdot d \cdot \sigma^2 \leq \frac{K^2 \gamma^3 TM^2 \chi \mu_{\max}}{\gamma TM} \cdot d \cdot \sigma^2$$

$$\leq \gamma^2 K^2 M \chi \mu_{\max} \cdot d \cdot \sigma^2$$

Hence, we prove the ergodic convergence rate.

$$\frac{1}{T} \sum_{t=1}^{T} \mathbb{E} \|\boldsymbol{\nabla} Q(\boldsymbol{x}_t)\|^2 \leq (2 + \mu_{\max} + \gamma KM \chi \mu_{\max}) \cdot \gamma K \cdot d \cdot \sigma^2 + d \cdot \sigma^2 + 2DK\sigma\sqrt{d} + K^2 D^2$$

□

**Theorem 5** (Convergence time complexity)**.** *Given a mini-batch size $|\xi|$, the number of gradients $M$ the server waits for before updating the model, and the total number of epochs $T$, the time complexity for convergence of Kardam is:*

$$\mathcal{O}\left( \frac{\mu_{\max}}{\sqrt{T \cdot |\xi| \cdot M}} + \frac{\chi \cdot \mu_{\max}}{T} + d \cdot \sigma^2 + 2DK\sigma\sqrt{d} + K^2 D^2 \right)$$

*where $\chi$ denotes a constant such that for all $\tau_{tl}$, the following inequality holds:*

$$\tau_{tl} \cdot \Lambda(\tau_{tl}) \leq \chi \qquad (4)$$

Theorem 5 highlights the relation between the staleness and the convergence time complexity. Furthermore, this time complexity is linearly dependent on the decay bound ($\chi$) and the maximum adaptive rate ($\mu_{max}$).

We now prove Theorem 5 by employing Theorem 4 along with Definition 4.

*Proof.* Substituting the value of $\gamma$ from Definition 4 in RHS of Theorem 4, we get the following.

$$\begin{aligned}(2 + \mu_{\max} + \gamma K M \chi \mu_{\max}) \cdot \gamma K \cdot d \cdot \sigma^2 + d \cdot \sigma^2 + 2DK\sigma\sqrt{d} + K^2 D^2 \\ = \mathcal{O}\left(\frac{\mu_{\max}}{\sqrt{T \cdot |\xi| \cdot M}} + \frac{\chi \cdot \mu_{\max}}{T} + d \cdot \sigma^2 + 2DK\sigma\sqrt{d} + K^2 D^2\right)\end{aligned}$$

Note that $\sigma = \mathcal{O}(1/\sqrt{|\xi|})$ (Assumption 2) and therefore the bound is also dependent on $n$.

□

**Remark 4** (Dampening functions comparison)**.** *Given two dampening functions* $\Lambda_1(\tau) = \frac{1}{1+\tau}$ *and* $\Lambda_2(\tau) = exp(-\alpha \sqrt[\beta]{\tau})$*, and the convergence time complexity from Theorem 5,* $\Lambda_2(\tau)$ *converges faster than* $\Lambda_1(\tau)$ *when* $\frac{\beta}{e} < \alpha \leq \frac{ln(\tau+1)}{\sqrt[\beta]{\tau}}$*.*

We also empirically highlight this remark by comparing these two functions in our main paper where DynSGD [4] employs $\Lambda_1$ and Kardam employs $\Lambda_2$.

*Proof.* From Inequality 4, we have the following for $\Lambda_1$ and $\Lambda_2$.

$$\chi_1 = \max_{\tau}\left\{\frac{\tau}{\tau+1}\right\}$$
$$\chi_2 = \max_{\tau}\left\{\tau \cdot exp(-\alpha \sqrt[\beta]{\tau})\right\}$$

The maximum value of $\{\tau \cdot exp(-\alpha \sqrt[\beta]{\tau})\}$ is $\left(\frac{\beta}{e\alpha}\right)^{\beta}$ when $\tau = \left(\frac{\beta}{\alpha}\right)^{\beta}$. We get that $\chi_1 \geq \chi_2$ when the following holds.

$$\frac{\tau}{\tau+1} \geq \left(\frac{\beta}{e\alpha}\right)^{\beta}$$

Hence, from the above inequality, we get the following.

$$\tau \geq \frac{1}{\left(\frac{e\alpha}{\beta}\right)^{\beta} - 1}$$

Note that since $\tau > 0$, we get $\left(\frac{e\alpha}{\beta}\right)^{\beta} > 1$ which leads to the following lower bound on $\alpha$.

$$\alpha > \frac{\beta}{e} \tag{5}$$

Furthermore, for the $\mu_{max}$ terms, we compare the values between the two dampening functions.

$$\mu_1 = \max_{\tau} \left\{ \frac{M}{\sum_{\mathbf{\Lambda}_t} \lambda \cdot |\mathcal{G}_{t\lambda}|} \right\} = \max_{\tau} \left\{ \frac{M}{\sum_{\tau} \frac{1}{\tau+1} \cdot |\mathcal{G}_{t\lambda}|} \right\}$$
$$\mu_2 = \max_{\tau} \left\{ \frac{M}{\sum_{\mathbf{\Lambda}_t} \lambda \cdot |\mathcal{G}_{t\lambda}|} \right\} = \max_{\tau} \left\{ \frac{M}{\sum_{\tau} exp(-\alpha \sqrt[\beta]{\tau}) \cdot |\mathcal{G}_{t\lambda}|} \right\}$$

Hence, for $\mu_1 \geq \mu_2$, we need to show that $\frac{1}{\tau+1} \leq exp(-\alpha \sqrt[\beta]{\tau})$, i.e., $\tau + 1 \geq exp(\alpha \sqrt[\beta]{\tau})$. The relation holds for any $\alpha$ with the upper bound as follows.

$$\alpha \leq \frac{ln(\tau+1)}{\sqrt[\beta]{\tau}} \tag{6}$$

From Inequalities 5 and 6, we get the following.

$$\frac{\beta}{e} < \alpha \leq \frac{ln(\tau+1)}{\sqrt[\beta]{\tau}}$$

One possible setting is $\beta \approx 1.85$ when $1 \leq \tau \leq 10$, $\beta \approx 3.1$ when $11 \leq \tau \leq 33$, and $\beta \approx 4$ when $34 \leq \tau \leq 75$. Given these values of $\beta$ and $\tau$, $\Lambda_2(\tau)$ has a smaller convergence time complexity (Theorem 5) than $\Lambda_1(\tau)$. Hence, $\Lambda_2(\tau)$ converges faster than $\Lambda_1(\tau)$.

□

# 3 Additional Experimental Results.

We also evaluate the performance of Kardam for image classification on the EMNIST dataset[2] consisting 814,255 examples of handwritten characters and digits (62 classes). We perform min-max scaling normalization as a pre-processing step resulting in 784 normalized input. We split the dataset into 697,932 training and 116,323 test examples and employ a base learning rate of $8 * 10^{-4}$ alongside a mini-batch of 100 examples if not stated otherwise.

[2] https://www.nist.gov/itl/iad/image-group/emnist-dataset

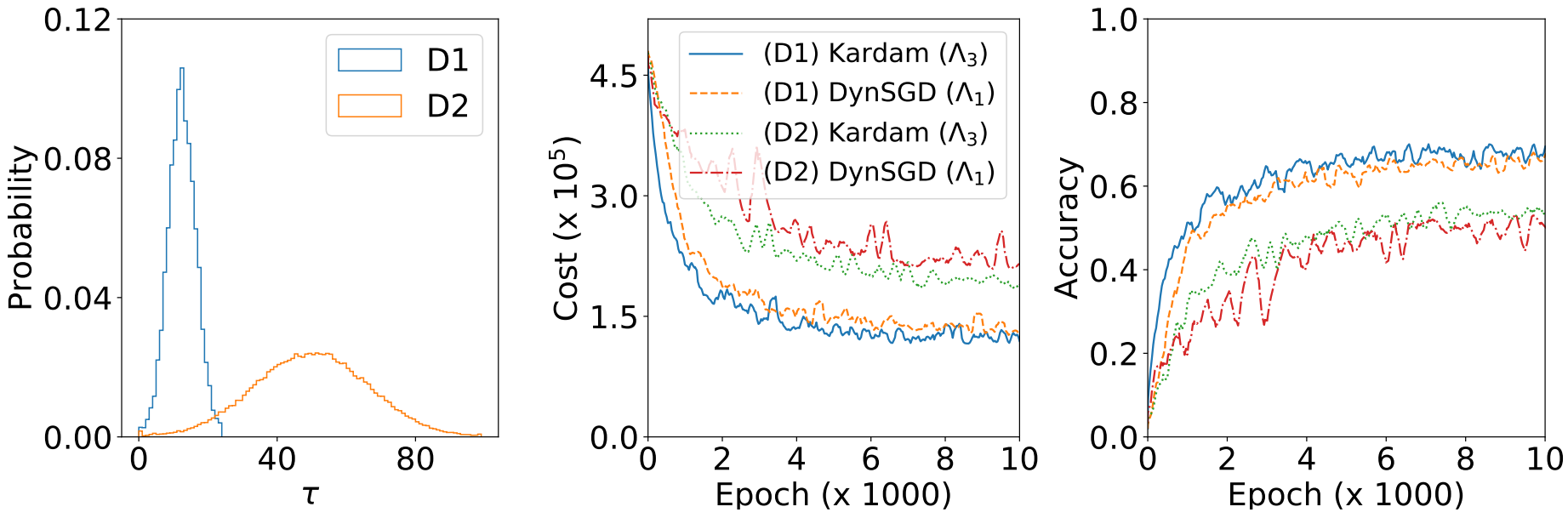

(a) Staleness distributions (b) Cost (c) Accuracy

**Figure 2: Impact of staleness for EMNIST.**

# References

[1] P. Blanchard, E. M. El Mhamdi, R. Guerraoui, and J. Stainer. Machine learning with adversaries: Byzantine tolerant gradient descent. In *NIPS*, pages 118–128, 2017.

[2] L. Bottou. Online learning and stochastic approximations. *Online learning in neural networks*, 17(9):142, 1998.

[3] M. J. Fischer, N. A. Lynch, and M. S. Paterson. Impossibility of distributed consensus with one faulty process. *JACM*, 32(2):374–382, 1985.

[4] J. Jiang, B. Cui, C. Zhang, and L. Yu. Heterogeneity-aware distributed parameter servers. In *SIGMOD*, pages 463–478, 2017.

[5] X. Lian, Y. Huang, Y. Li, and J. Liu. Asynchronous parallel stochastic gradient for nonconvex optimization. In *NIPS*, pages 2737–2745, 2015.

[6] R. Zhang, S. Zheng, and J. T. Kwok. Asynchronous distributed semi-stochastic gradient optimization. In *AAAI*, pages 2323–2329, 2016.

[7] W. Zhang, S. Gupta, X. Lian, and J. Liu. Staleness-aware async-sgd for distributed deep learning. In *IJCAI*, pages 2350–2356, 2016.